\documentclass{article} 
\usepackage{template/iclr2026_conference}
\usepackage{times}

\usepackage{amsmath,amsfonts,bm}

\def\eqref#1{equation~\ref{#1}}

\def\1{\bm{1}}

\DeclareMathAlphabet{\mathsfit}{\encodingdefault}{\sfdefault}{m}{sl}
\SetMathAlphabet{\mathsfit}{bold}{\encodingdefault}{\sfdefault}{bx}{n}

\usepackage{amsmath}
\usepackage{amssymb}
\usepackage{amsthm}

\theoremstyle{remark}

\usepackage{graphicx}
\usepackage{booktabs}
\usepackage{multirow}
\usepackage{adjustbox}
\usepackage{tikz}
\usetikzlibrary{arrows.meta,positioning,calc}
\usepackage{xcolor}
\usepackage{wrapfig}
\usepackage{enumitem}

\usepackage{xspace}

\newcommand{\method}{MapLightning\xspace}

\definecolor{iclrblue}{rgb}{0.21,0.49,0.74}

\definecolor{textgreen}{rgb}{0.4980392156862745, 0.788235294117647, 0.4980392156862745}
\definecolor{ourmediumblue}{rgb}{0.21568627450980393,0.49411764705882355,0.7215686274509804}
\definecolor{ourmediumred}{rgb}{0.8941176470588236,0.10196078431372549,0.10980392156862745}
\definecolor{ourmediumgreen}{rgb}{0.30196078431372547,0.6862745098039216,0.2901960784313726}

\definecolor{ourlightred}{rgb}{0.984313725490196, 0.5019607843137255, 0.4470588235294118}
\definecolor{ourlightblue}{rgb}{0.5019607843137255, 0.6941176470588235, 0.8274509803921568}
\definecolor{ourlightcyan}{rgb}{0.5529411764705883, 0.8274509803921568, 0.7803921568627451}

\newcommand{\txtgreen}[1]{\textcolor{textgreen}{#1}}

\newcommand{\txtred}[1]{\textcolor{ourmediumred}{#1}}

\definecolor{RdYlGn1}{HTML}{d73027}\definecolor{RdYlGn2}{HTML}{f46d43}\definecolor{RdYlGn3}{HTML}{fdae61}
\definecolor{RdYlGn4}{HTML}{fee08b}\definecolor{RdYlGn5}{HTML}{ffffbf}\definecolor{RdYlGn6}{HTML}{d9ef8b}
\definecolor{RdYlGn7}{HTML}{a6d96a}\definecolor{RdYlGn8}{HTML}{66bd63}\definecolor{RdYlGn9}{HTML}{1a9850}
\definecolor{RdYlBu1}{HTML}{d73027}\definecolor{RdYlBu2}{HTML}{f46d43}\definecolor{RdYlBu3}{HTML}{fdae61}
\definecolor{RdYlBu4}{HTML}{fee090}\definecolor{RdYlBu5}{HTML}{ffffbf}\definecolor{RdYlBu6}{HTML}{e0f3f8}
\definecolor{RdYlBu7}{HTML}{abd9e9}\definecolor{RdYlBu8}{HTML}{74add1}\definecolor{RdYlBu9}{HTML}{4575b4}
\definecolor{Spectral1}{HTML}{d53e4f}\definecolor{Spectral2}{HTML}{f46d43}\definecolor{Spectral3}{HTML}{fdae61}
\definecolor{Spectral4}{HTML}{fee08b}\definecolor{Spectral5}{HTML}{ffffbf}\definecolor{Spectral6}{HTML}{e6f598}
\definecolor{Spectral7}{HTML}{abdda4}\definecolor{Spectral8}{HTML}{66c2a5}\definecolor{Spectral9}{HTML}{3288bd}
\definecolor{PRGn1}{HTML}{762a83}\definecolor{PRGn2}{HTML}{9970ab}\definecolor{PRGn3}{HTML}{c2a5cf}
\definecolor{PRGn4}{HTML}{e7d4e8}\definecolor{PRGn5}{HTML}{f7f7f7}\definecolor{PRGn6}{HTML}{d9f0d3}
\definecolor{PRGn7}{HTML}{a6dba0}\definecolor{PRGn8}{HTML}{5aae61}\definecolor{PRGn9}{HTML}{1b7837}
\definecolor{RdBu1}{HTML}{b2182b}\definecolor{RdBu2}{HTML}{d6604d}\definecolor{RdBu3}{HTML}{f4a582}
\definecolor{RdBu4}{HTML}{fddbc7}\definecolor{RdBu5}{HTML}{f7f7f7}\definecolor{RdBu6}{HTML}{d1e5f0}
\definecolor{RdBu7}{HTML}{92c5de}\definecolor{RdBu8}{HTML}{4393c3}\definecolor{RdBu9}{HTML}{2166ac}
\definecolor{RdGy1}{HTML}{b2182b}\definecolor{RdGy2}{HTML}{d6604d}\definecolor{RdGy3}{HTML}{f4a582}
\definecolor{RdGy4}{HTML}{fddbc7}\definecolor{RdGy5}{HTML}{ffffff}\definecolor{RdGy6}{HTML}{e0e0e0}
\definecolor{RdGy7}{HTML}{bababa}\definecolor{RdGy8}{HTML}{878787}\definecolor{RdGy9}{HTML}{4d4d4d}
\definecolor{PiYG1}{HTML}{c51b7d}\definecolor{PiYG2}{HTML}{de77ae}\definecolor{PiYG3}{HTML}{f1b6da}
\definecolor{PiYG4}{HTML}{fde0ef}\definecolor{PiYG5}{HTML}{f7f7f7}\definecolor{PiYG6}{HTML}{e6f5d0}
\definecolor{PiYG7}{HTML}{b8e186}\definecolor{PiYG8}{HTML}{7fbc41}\definecolor{PiYG9}{HTML}{4d9221}
\definecolor{BrBG1}{HTML}{8c510a}\definecolor{BrBG2}{HTML}{bf812d}\definecolor{BrBG3}{HTML}{dfc27d}
\definecolor{BrBG4}{HTML}{f6e8c3}\definecolor{BrBG5}{HTML}{f5f5f5}\definecolor{BrBG6}{HTML}{c7eae5}
\definecolor{BrBG7}{HTML}{80cdc1}\definecolor{BrBG8}{HTML}{35978f}\definecolor{BrBG9}{HTML}{01665e}

\usepackage[pagebackref,breaklinks,colorlinks,allcolors=iclrblue]{hyperref}
\usepackage{url}

\let\cite\citep

\title{MapLightning: Online Vectorized HD Map\\ Construction with 1D Map Tokens}

\author{Shen Zheng \And Anurag Ghosh \And Mani Ramanagopal \And Srinivasa Narasimhan
\AND \makebox[\textwidth][c]{\raisebox{0.7em}[0pt][0pt]{\mdseries Carnegie Mellon University}}}

\iclrfinalcopy 
\begin{document}

\maketitle

\vspace{-1.7em}

\begin{abstract}

Online vectorized HD map construction is essential for scaling safe autonomous driving and requires accurate, real-time inference. Prior methods typically rely on dense bird's-eye-view (BEV) grids as the intermediate representation. We propose \textit{\method{}}, which replaces the dense BEV grid with a compact set of 1D learnable map tokens. To construct map tokens from image features, we choose self-attention over vanilla cross-attention because it enables joint interactions and contextual aggregation among image and map tokens. Our transformer-based mapper concatenates map and image tokens, applies full self-attention, discards the image tokens, and retains the updated map tokens for decoding. This design offers three advantages. First, our representation is efficient, using fewer tokens, consuming less memory, and running faster. Second, the lightweight design allows the map decoder to use full rather than deformable cross-attention for better global context. Third, unlike BEV-based methods, our network does not use camera projection parameters, making it robust to camera-extrinsic perturbations. \method{} uses up to 16.7$\times$ fewer intermediate tokens than dense BEV-based methods and achieves state-of-the-art accuracy and efficiency on nuScenes and Argoverse~2. Its lightweight variant surpasses MapTRv2 by +10.1 mAP on nuScenes and +16.2 mAP on Argoverse~2, while delivering 1.73$\times$ faster inference (40+ FPS) with 53\% less memory. We further show improvements on uncertainty-aware map construction and downstream trajectory prediction. Code and models will be released.


\end{abstract}


\section{Introduction}
\label{sec:intro}

High-definition (HD) maps provide centimeter-level vectorized representations of static road elements such as lane dividers, pedestrian crossings, and road boundaries. These elements encode road topology and driving constraints essential for autonomous navigation. Traditionally, HD maps are constructed offline using SLAM- or SfM-based pipelines~\cite{zhang2014loam,shan2018lego,zhanabatyrova2024structure}, requiring expensive annotation and maintenance. 

Recently, online vectorized HD map construction~\cite{liao2022maptr,yuan2024streammapnet,chen2024maptracker,liao2025maptrv2,dong2025damap,monninger2026augmapnet} has gained popularity by building local maps at runtime from onboard sensors, reducing reliance on costly offline map construction and maintenance. Online vectorized HD map construction methods typically follow a three-stage pipeline: an image encoder extracts features from surround-view images, a bird’s-eye-view (BEV) projection module such as LSS~\cite{philion2020lift} or BEVFormer~\cite{li2024bevformer} projects them into a dense BEV grid, and a DETR-style~\cite{carion2020end} decoder with deformable attention~\cite{zhu2020deformable} predicts map elements as ordered point sets. 



The BEV-based design is effective for dense spatial perception tasks like semantic segmentation~\cite{philion2020lift,zhou2022cross} and 3D occupancy prediction~\cite{yu2023flashocc,hou2024fastocc}, where dense spatial prediction naturally aligns with the BEV grid. However, it is less suited for vectorized map prediction, where the output is a sparse set of vectorized map elements that occupy only \textbf{4.27\%} of the rasterized perception range on nuScenes~\cite{caesar2020nuscenes} and \textbf{2.65\%} on Argoverse 2~\cite{wilson2023argoverse}. 

In this paper, we propose to use 1D learnable tokens, which we call \emph{map tokens}, as the intermediate scene representation. This design offers three key advantages. \textbf{(i)} Our representation uses fewer intermediate tokens, consumes less memory, and runs faster. \textbf{(ii)} The lightweight design allows the map decoder to use full attention instead of deformable attention, giving each query access to the entire representation for better global context and long-range dependencies. \textbf{(iii)} Unlike BEV-based methods, ours does not use camera projection parameters, making it robust to extrinsic perturbations.

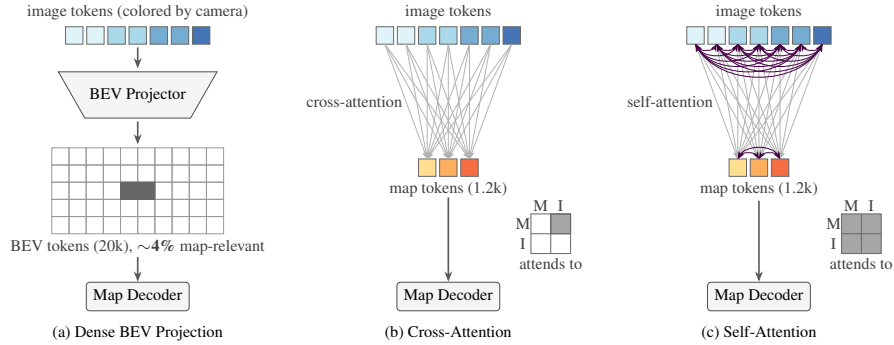
\begin{figure}[t]
\centering
\adjustbox{max width=0.85\linewidth}{%
\begin{tikzpicture}[
  font=\scriptsize,
  tok/.style={draw=black!60, line width=0.3pt, minimum size=7.5pt, inner sep=0pt, outer sep=0pt},
  proc/.style={draw=black!70, fill=black!4, rounded corners=1.5pt, inner sep=3pt, align=center},
  dec/.style={draw=black!70, fill=black!4, rounded corners=1.5pt, inner sep=3pt},
  flow/.style={-{Stealth[length=4.5pt]}, black!65, line width=0.7pt},
  att/.style={-{Stealth[length=3pt]}, black!55, line width=0.45pt},
  attb/.style={{Stealth[length=3pt]}-{Stealth[length=3pt]}, black!55, line width=0.45pt},
  lab/.style={text=black!75, align=center},
]

\def\imgcolors{{"RdYlBu6","RdYlBu6","RdYlBu7","RdYlBu7","RdYlBu8","RdYlBu8","RdYlBu9","RdYlBu9"}}
\def\scenecolors{{"RdYlBu4","RdYlBu3","RdYlBu2"}}

\begin{scope}[shift={(0,0)}]
  \foreach \i in {0,...,6}{
    \pgfmathsetmacro{\c}{\imgcolors[\i]}
    \node[tok, fill=\c] at (1.29+0.32*\i, 4.0) {};
  }
  \node[lab] at (2.25,4.35) {image tokens (colored by camera)};
  
  \draw[flow] (2.25,3.82) -- (2.25,3.50);
  \draw[black!70, fill=black!4] (1.05,3.45) -- (3.45,3.45) -- (3.05,2.75) -- (1.45,2.75) -- cycle;
  \node at (2.25,3.10) {BEV Projector};
        
    \draw[flow] (2.25,2.70) -- (2.25,2.35);
    \foreach \col in {0,...,9}{
      \foreach \row in {0,...,4}{
        \draw[black!40, line width=0.3pt]
          (0.95+0.26*\col, 1.00+0.26*\row)
          rectangle ++(0.26,0.26);
      }
    }

\draw[black!60, fill=black!60, line width=0.3pt]
  (1.99,1.52) rectangle ++(0.26,0.26);

\draw[black!60, fill=black!60, line width=0.3pt]
  (2.25,1.52) rectangle ++(0.26,0.26);
  
  
  \node[lab] at (2.25,0.80) {BEV tokens (20k), ${\sim}$\textbf{4\%} map-relevant};
    
  \node[dec] (decA) at (2.25,0.05) {Map Decoder};
  \draw[flow] (2.25,0.65) -- (decA.north);
  \node at (2.25,-0.5) {(a) Dense BEV Projection};
\end{scope}

\begin{scope}[shift={(4.7,0)}]
  \foreach \i in {0,...,6}{
    \pgfmathsetmacro{\c}{\imgcolors[\i]}
    \node[tok, fill=\c] (bimg\i) at (1.29+0.32*\i, 4.0) {};
  }
  \node[lab] at (2.25,4.35) {image tokens};

  \foreach \i in {0,...,2}{
    \pgfmathsetmacro{\c}{\scenecolors[\i]}
    \node[tok, fill=\c] (bscn\i) at (1.93+0.32*\i, 2.0) {};
  }
  \node[lab] at (2.25,1.67) {map tokens (1.2k)};

  \foreach \i in {0,...,6}{
    \foreach \j in {0,...,2}{
      \draw[-{Stealth[length=3pt]}, black!30, line width=0.35pt]
        (bimg\i.south) -- (bscn\j.north);
    }
  }

  \node[lab, anchor=west] at (-0.02,3.0) {cross-attention};

  \begin{scope}[shift={(3.5,0.7)}]
    \draw[black!60, fill=black!35] (0.3,0.3) rectangle (0.6,0.6);
    \draw[black!60] (0,0) rectangle (0.6,0.6);
    \draw[black!60] (0.3,0) -- (0.3,0.6);
    \draw[black!60] (0,0.3) -- (0.6,0.3);
    \node[lab] at (-0.12,0.45) {M};
    \node[lab] at (-0.12,0.15) {I};
    \node[lab] at (0.15,0.72) {M};
    \node[lab] at (0.45,0.72) {I};
    \node[lab] at (0.3,-0.15) {attends to};
  \end{scope}

  \node[dec] (decB) at (2.25,0.05) {Map Decoder};
  \draw[flow] (2.25,1.55) -- (decB.north);

  \node at (2.25,-0.5) {(b) Cross-Attention};
\end{scope}

\begin{scope}[shift={(9.4,0)}]
  \foreach \i in {0,...,6}{
    \pgfmathsetmacro{\c}{\imgcolors[\i]}
    \node[tok, fill=\c] (cimg\i) at (1.29+0.32*\i, 4.0) {};
  }
  \node[lab] at (2.25,4.35) {image tokens};

  \foreach \i in {0,...,2}{
    \pgfmathsetmacro{\c}{\scenecolors[\i]}
    \node[tok, fill=\c] (crd\i) at (1.93+0.32*\i, 2.0) {};
  }
  \node[lab] at (2.25,1.67) {map tokens (1.2k)};

  \foreach \i in {0,...,6}{
    \foreach \j in {0,...,2}{
      \draw[{Stealth[length=3pt]}-{Stealth[length=3pt]}, black!30, line width=0.35pt]
        (cimg\i.south) -- (crd\j.north);
    }
  }

  \foreach \i in {0,...,6}{
    \foreach \j in {0,...,6}{
      \ifnum\i<\j
        \draw[{Stealth[length=3pt]}-{Stealth[length=3pt]}, violet!55!black, line width=0.35pt]
          (cimg\i.south) to[bend right=60] (cimg\j.south);   
      \fi
    }
  }

  \foreach \i in {0,...,2}{
    \foreach \j in {0,...,2}{
      \ifnum\i<\j
        \draw[{Stealth[length=3pt]}-{Stealth[length=3pt]}, violet!55!black, line width=0.35pt]
          (crd\i.north) to[bend left=60] (crd\j.north);   
      \fi
    }
  }

  \node[lab, anchor=west] at (0.13,3.0) {self-attention};

  \begin{scope}[shift={(3.5,0.7)}]
    \draw[black!60, fill=black!35] (0,0) rectangle (0.6,0.6);
    \draw[black!60] (0.3,0) -- (0.3,0.6);
    \draw[black!60] (0,0.3) -- (0.6,0.3);
    \node[lab] at (-0.12,0.45) {M};
    \node[lab] at (-0.12,0.15) {I};
    \node[lab] at (0.15,0.72) {M};
    \node[lab] at (0.45,0.72) {I};
    \node[lab] at (0.3,-0.15) {attends to};
  \end{scope}

  \node[dec] (decC) at (2.25,0.05) {Map Decoder};
  \draw[flow] (2.25,1.50) -- (decC.north);

  \node at (2.25,-0.5) {(c) Self-Attention};
\end{scope}

\end{tikzpicture}%
}
\vspace{-1em}
\caption{\textbf{Aggregating multi-view image tokens into a scene representation.} \textbf{(a)}~BEV-based methods~\cite{liao2022maptr,liao2025maptrv2} geometrically project image features onto a dense grid of fixed BEV cells, few of which contain map elements. \textbf{(b)}~Vanilla cross-attention compresses image tokens into compact map tokens, but image tokens serve only as fixed keys and values, with no image-to-image or map-to-map interaction. \textbf{(c)}~Self-attention (ours, Section~\ref{subsec:map_mapper}) attends over the concatenated map and image tokens, refining all tokens jointly. Gray and violet edges denote image--map and intra-group attention, respectively. In the masks, rows are queries, columns are keys/values, shaded cells mark allowed attention, and $\mathrm{M}$/$\mathrm{I}$ denote map/image tokens. See Table~\ref{tab:mechanism} for comparison.}
\label{fig:aggregation}
\vspace{-1.5em}
\end{figure}
\begin{figure}[t]
\centering
 \vspace{-0.2cm}
\includegraphics[width=0.9\linewidth]{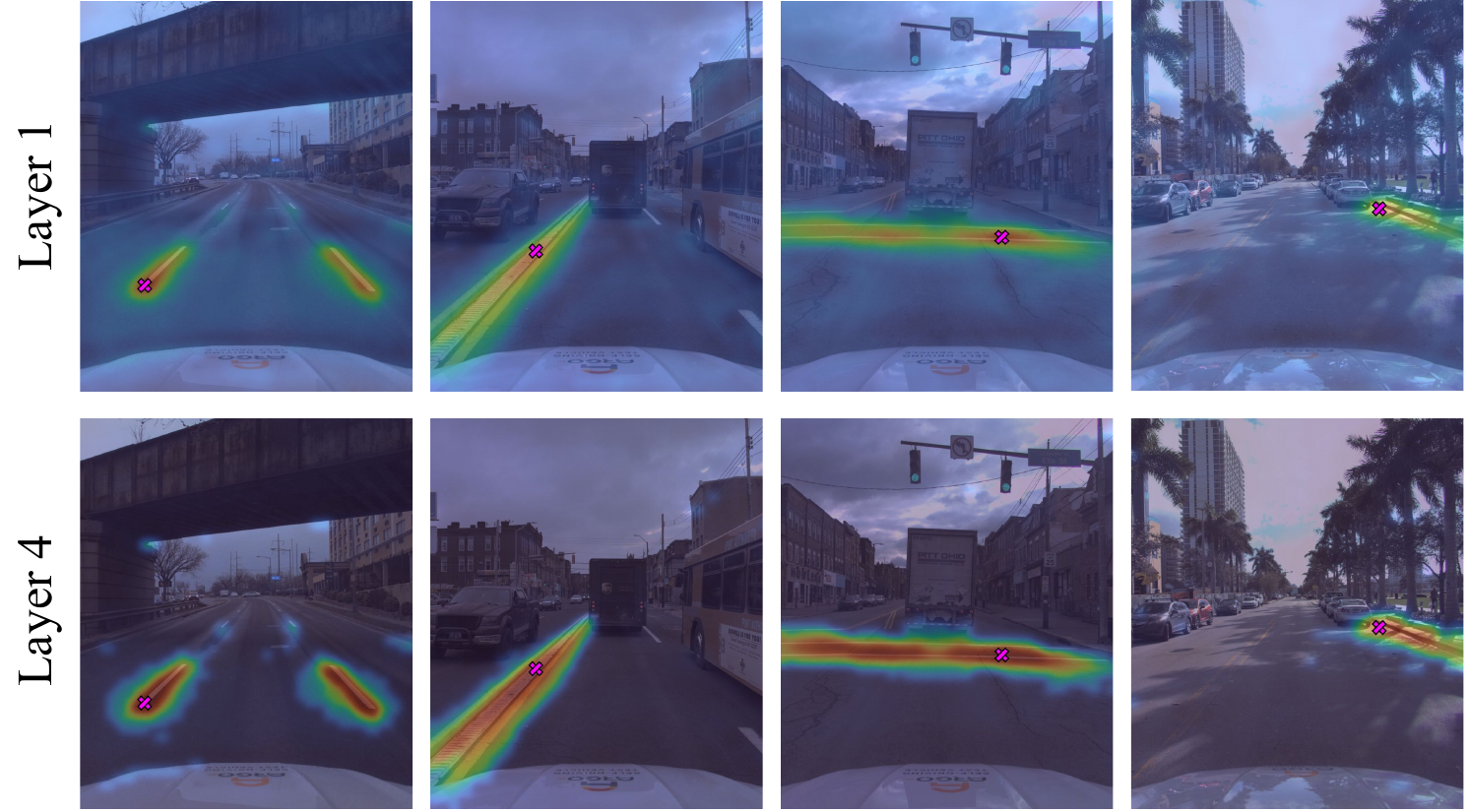}
 \vspace{-0.3cm}
\caption{\textbf{Image-to-image attention in the self-attention mapper.} We select one image token as the query (magenta cross) and visualize its attention scores over the other image tokens, which serve as keys (red: high, blue: low), at the first and last ($L=4$) mapper layers. From left to right: (1)~a query on a lane divider attends along that divider and to the parallel divider across the lane; (2)~a query on a lane divider attends along the divider's full extent toward the horizon; (3)~a query on a pedestrian crossing attends across the crossing's full width; and (4)~a query on a road boundary attends along the curb. From Layer~1 to Layer~4, responses on these structures become stronger and more complete, while diffuse background responses (e.g., the sky in column 4) fade, illustrating progressive refinement of the image tokens. Such image-to-image interaction is absent in cross-attention, where image tokens serve only as fixed keys and values. Best viewed zoomed in.}

\label{fig:attn_refinement}
\end{figure}

Constructing map tokens requires compressing a large number of multi-view image tokens into a much smaller set, and the choice of compression mechanism is critical. Specifically, we concatenate map and image tokens and apply attention over the joint sequence. As shown in Figure~\ref{fig:aggregation}, we adopt self-attention instead of vanilla cross-attention, since vanilla cross-attention only lets map tokens read from image tokens, whereas self-attention also allows explicit image-to-image and map-to-map token interactions for richer contextual aggregation. The image-to-image interactions are particularly useful for refining visual context: as illustrated in Figure~\ref{fig:attn_refinement}, self-attention progressively concentrates on map-relevant structures across layers. After the final mapper layer, the image tokens are discarded, leaving compact map tokens that retain the aggregated map-relevant information. These map tokens serve as the intermediate representation for the decoder. In the decoder, cross-attention is used because the map queries act as output slots that read from the fixed map-token representation, similar to decoding objects~\cite{carion2020end} or categories~\cite{strudel2021segmenter}.

We name our framework \textit{\method{}} for its fast inference, token-efficient design, and high accuracy. By replacing the dense BEV grid with a compact set of tokens (up to \textbf{16.7$\times$} fewer), \method{} achieves state-of-the-art accuracy and efficiency on nuScenes and Argoverse 2. With a lightweight MobileNetV3-Large backbone, it surpasses MapTRv2~\cite{liao2025maptrv2} by \textbf{\txtgreen{+10.1}} and \textbf{\txtgreen{+16.2}} mAP on nuScenes and Argoverse 2, running \textbf{1.73$\times$} faster (\textbf{40+ FPS}) with \textbf{53\%} less memory.



\noindent In summary, the main contributions of this paper are:
\begin{itemize}[leftmargin=*, topsep=2pt, itemsep=2pt, parsep=0pt]
   \item We propose \method{}, the first online vectorized HD map construction framework that replaces dense BEV projection with attention-based mapping into compact 1D map tokens, yielding higher efficiency, better global context, and robustness to extrinsic perturbations.
    \item We compare cross-attention and self-attention for constructing map tokens from image features, and show that self-attention is more effective for online vectorized HD map construction.
    \item \method{} outperforms prior methods on online vectorized HD map construction, uncertainty-aware map construction, downstream trajectory prediction, and 3D object detection.
\end{itemize}
\vspace{-1em}
\section{Related Work}
\label{sec:related}

\subsection{Online Vectorized HD Map Construction} 

Most online vectorized HD map construction methods follow a MapTR-style~\cite{liao2022maptr} framework with dense BEV representations and a DETR-style decoder~\cite{liu2023vectormapnet,liao2022maptr,liao2025maptrv2}. Follow-up works refine the decoder, supervision, and temporal modeling within this paradigm~\cite{yuan2024streammapnet,chen2024maptracker, dong2025damap,monninger2026augmapnet}. 

Our approach differs from prior online vectorized HD map construction methods in two key aspects: \textbf{(i)} Instead of following the conventional dense BEV paradigm, we compress image tokens into compact 1D map tokens using attention, achieving higher efficiency and accuracy. \textbf{(ii)} Unlike many previous methods that focus on map construction alone, we show that our representation benefits downstream trajectory prediction and generalizes to 3D object detection.

\subsection{Beyond BEV-Based Representation} 

BEV representations have been widely adopted for dense spatial perception. For example, Lift-Splat-Shoot~\cite{philion2020lift} and Cross-View Transformers~\cite{zhou2022cross} construct BEV features for semantic segmentation, whereas FlashOcc~\cite{yu2023flashocc} and FastOcc~\cite{hou2024fastocc} extend this representation to 3D occupancy prediction.

Works on sparse prediction have explored removing BEV as an intermediate representation by aggregating image features into task queries. In 3D object detection, SparseBEV~\cite{liu2023sparsebev} achieves performance comparable to dense-BEV methods like BEVFormer v2~\cite{yang2023bevformer}. In online vectorized HD map construction, SparseMeXt~\cite{jiang2025sparsemext} follows a similar direction but underperforms dense-BEV methods like MapTRv2~\cite{liao2025maptrv2} in comparable settings.

Unlike prior methods that either retain BEV-based feature projection or drop the intermediate representation, we use 1D learnable map tokens as a compact and performant intermediate representation.

\subsection{Compressing Tokens via Attention Mechanisms} 

Learned query tokens are a common pattern for pooling a large set of input tokens via cross-attention. Queries can decode specific outputs, such as object predictions~\cite{carion2020end} or slots that explain an image~\cite{locatello2020object,vani2024sparo}. Queries can also compress tokens into a smaller bottleneck for further processing, as in Perceiver's latents~\cite{jaegle2021perceiver,jaegleperceiverio}, Flamingo's resampler~\cite{alayrac2022flamingo}, and Q-Former~\cite{li2023blip}.

Learned tokens can also aggregate information through self-attention by being concatenated with input tokens and jointly updated within the same sequence, which most closely resembles our design. Such tokens have been explored in general representation learning and sequence modeling, including the \texttt{[CLS]} token in BERT~\cite{devlin2019bert}, memory tokens~\cite{burtsev2020memory}, the \texttt{[CLS]} token in ViT~\cite{dosovitskiy2021image}, gist tokens~\cite{mu2023learning}, summary vectors~\cite{chevalier2023adapting}, register tokens in Registers~\cite{darcet2024vision}, and planning tokens in Flex~\cite{yang2025towards}. They have also been used in dense or generative tasks, such as latent tokens in LVSM~\cite{jin2024lvsm} and light tokens in RelitLRM~\cite{zhang2025relitlrm}. 

Unlike previous works targeting dense or general-purpose perception, we apply attention-based token compression to the sparse task of online vectorized HD map construction with a tailored mapper.

\section{Method}
\label{sec:method}

\usetikzlibrary{shapes.geometric}

\begin{figure}[t]
\centering
\adjustbox{max width=\linewidth}{%
\begin{tikzpicture}[
  font=\scriptsize,
  tok/.style={draw=black!60, line width=0.3pt, minimum size=7.5pt, inner sep=0pt, outer sep=0pt},
  pict/.style={draw=black!60, line width=0.3pt, rounded corners=0.8pt, minimum width=7pt, minimum height=5.2pt, inner sep=0pt, outer sep=0pt},
  proc/.style={draw=black!70, fill=black!4, rounded corners=1.5pt, inner sep=3pt, align=center, minimum height=0.55cm},
  dec/.style={draw=black!70, fill=black!4, rounded corners=1.5pt, inner sep=3pt, align=center, minimum height=0.55cm},
  flow/.style={-{Stealth[length=4.5pt]}, black!65, line width=0.7pt},
  att/.style={-{Stealth[length=3pt]}, black!55, line width=0.45pt},
  lab/.style={text=black!75, align=center},
]

\colorlet{camA}{RdYlBu6}
\colorlet{camB}{RdYlBu6!50!RdYlBu7}
\colorlet{camC}{RdYlBu7}
\colorlet{camD}{RdYlBu7!50!RdYlBu8}
\colorlet{camE}{RdYlBu8}
\colorlet{camF}{RdYlBu9}
\def\camcolors{{"camA","camB","camC","camD","camE","camF"}}
\def\scenecolors{{"RdYlBu4","RdYlBu3","RdYlBu2"}}

\definecolor{mapblue}{RGB}{0,0,255}
\definecolor{maporange}{RGB}{255,165,0}
\definecolor{mapgreen}{RGB}{0,128,0}

\foreach \l in {0,...,3}{
  \pgfmathsetmacro{\dx}{0.09*\l}
  \pgfmathsetmacro{\dy}{-0.09*\l}
  \draw[black!50, line width=0.3pt, fill=white, rounded corners=1pt]
    (-0.3+\dx,3.4+\dy) rectangle (1.45+\dx,2.0+\dy);
}

\draw[black!70, densely dotted, line width=0.5pt]
  (0.825,2.68) -- (0.825,2.82);   

\draw[black!70, densely dotted, line width=0.5pt]
  (0.69,2.56) -- (0.48,2.63);     

\draw[black!70, densely dotted, line width=0.5pt]
  (0.69,2.34) -- (0.50,2.29);     

\draw[black!70, densely dotted, line width=0.5pt]
  (0.96,2.56) -- (1.15,2.61);     

\draw[black!70, densely dotted, line width=0.5pt]
  (0.96,2.34) -- (1.15,2.29);     

\draw[black!70, densely dotted, line width=0.5pt]
  (0.825,2.22) -- (0.825,2.08);   

\def\camw{0.18}
\def\camh{0.13}

\pgfmathsetmacro{\fTopY}{2.94-\camh + (2*\camh)/2}
\def\fTopHalf{0.035}

\begin{scope}
  \clip[rounded corners=0.5pt] (0.825-\camw,2.94-\camh) rectangle (0.825+\camw,2.94+\camh);
  \fill[black!30]
    (0.825-\camw,2.94-\camh) --
    (0.825+\camw,2.94-\camh) --
    (0.825+\fTopHalf,\fTopY) --
    (0.825-\fTopHalf,\fTopY) -- cycle;
  \draw[black!35, line width=0.22pt]
    (0.825-\camw,2.94-\camh) -- (0.825-\fTopHalf,\fTopY);
  \draw[black!35, line width=0.22pt]
    (0.825+\camw,2.94-\camh) -- (0.825+\fTopHalf,\fTopY);
\end{scope}
\draw[black!60, line width=0.25pt, rounded corners=0.5pt]
  (0.825-\camw,2.94-\camh) rectangle (0.825+\camw,2.94+\camh);

\pgfmathsetmacro{\flyL}{2.65-\camh + 0.33*(2*\camh)}
\pgfmathsetmacro{\flyR}{2.65-\camh + 0.50*(2*\camh)}
\begin{scope}
  \clip[rounded corners=0.5pt] (0.33-\camw,2.65-\camh) rectangle (0.33+\camw,2.65+\camh);
  \fill[black!30]
    (0.33-\camw,2.65-\camh) --
    (0.33+\camw,2.65-\camh) --
    (0.33+\camw,\flyR) --
    (0.33-\camw,\flyL) -- cycle;
  \draw[black!35, line width=0.22pt]
    (0.33-\camw,\flyL) -- (0.33+\camw,\flyR);
\end{scope}
\draw[black!60, line width=0.25pt, rounded corners=0.5pt]
  (0.33-\camw,2.65-\camh) rectangle (0.33+\camw,2.65+\camh);

\pgfmathsetmacro{\rlyL}{2.25-\camh + 0.50*(2*\camh)}
\pgfmathsetmacro{\rlyR}{2.25-\camh + 0.33*(2*\camh)}
\begin{scope}
  \clip[rounded corners=0.5pt] (0.33-\camw,2.25-\camh) rectangle (0.33+\camw,2.25+\camh);
  \fill[black!30]
    (0.33-\camw,2.25-\camh) --
    (0.33+\camw,2.25-\camh) --
    (0.33+\camw,\rlyR) --
    (0.33-\camw,\rlyL) -- cycle;
  \draw[black!35, line width=0.22pt]
    (0.33-\camw,\rlyL) -- (0.33+\camw,\rlyR);
\end{scope}
\draw[black!60, line width=0.25pt, rounded corners=0.5pt]
  (0.33-\camw,2.25-\camh) rectangle (0.33+\camw,2.25+\camh);

\pgfmathsetmacro{\fryL}{2.65-\camh + 0.50*(2*\camh)}
\pgfmathsetmacro{\fryR}{2.65-\camh + 0.33*(2*\camh)}
\begin{scope}
  \clip[rounded corners=0.5pt] (1.32-\camw,2.65-\camh) rectangle (1.32+\camw,2.65+\camh);
  \fill[black!30]
    (1.32-\camw,2.65-\camh) --
    (1.32+\camw,2.65-\camh) --
    (1.32+\camw,\fryR) --
    (1.32-\camw,\fryL) -- cycle;
  \draw[black!35, line width=0.22pt]
    (1.32-\camw,\fryL) -- (1.32+\camw,\fryR);
\end{scope}
\draw[black!60, line width=0.25pt, rounded corners=0.5pt]
  (1.32-\camw,2.65-\camh) rectangle (1.32+\camw,2.65+\camh);

\pgfmathsetmacro{\rryL}{2.25-\camh + 0.33*(2*\camh)}
\pgfmathsetmacro{\rryR}{2.25-\camh + 0.50*(2*\camh)}
\begin{scope}
  \clip[rounded corners=0.5pt] (1.32-\camw,2.25-\camh) rectangle (1.32+\camw,2.25+\camh);
  \fill[black!30]
    (1.32-\camw,2.25-\camh) --
    (1.32+\camw,2.25-\camh) --
    (1.32+\camw,\rryR) --
    (1.32-\camw,\rryL) -- cycle;
  \draw[black!35, line width=0.22pt]
    (1.32-\camw,\rryL) -- (1.32+\camw,\rryR);
\end{scope}
\draw[black!60, line width=0.25pt, rounded corners=0.5pt]
  (1.32-\camw,2.25-\camh) rectangle (1.32+\camw,2.25+\camh);

\pgfmathsetmacro{\rTopY}{1.96-\camh + (2*\camh)/2}
\def\rTopHalf{0.035}

\begin{scope}
  \clip[rounded corners=0.5pt] (0.825-\camw,1.96-\camh) rectangle (0.825+\camw,1.96+\camh);
  \fill[black!30]
    (0.825-\camw,1.96-\camh) --
    (0.825+\camw,1.96-\camh) --
    (0.825+\rTopHalf,\rTopY) --
    (0.825-\rTopHalf,\rTopY) -- cycle;
  \draw[black!35, line width=0.22pt]
    (0.825-\camw,1.96-\camh) -- (0.825-\rTopHalf,\rTopY);
  \draw[black!35, line width=0.22pt]
    (0.825+\camw,1.96-\camh) -- (0.825+\rTopHalf,\rTopY);
\end{scope}
\draw[black!60, line width=0.25pt, rounded corners=0.5pt]
  (0.825-\camw,1.96-\camh) rectangle (0.825+\camw,1.96+\camh);

\begin{scope}[shift={(0.825,2.45)}, scale=0.32]
  \draw[black!70, line width=0.4pt, fill=white, rounded corners=2pt]
    (-0.4,-0.7) rectangle (0.4,0.7);
  \draw[black!70, line width=0.4pt, fill=black!8, rounded corners=1.5pt]
    (-0.28,-0.35) rectangle (0.28,0.35);
  \foreach \wx/\wy in {-0.42/0.45, 0.42/0.45, -0.42/-0.45, 0.42/-0.45}{
    \draw[black!70, line width=0.3pt, fill=black!60]
      (\wx-0.06,\wy-0.14) rectangle (\wx+0.06,\wy+0.14);
  }
\end{scope}

\coordinate (vidR) at (1.6, 2.66);

\node[lab] at (0.575,3.6) {multi-view video};
\draw[att] (-0.45,3.4) -- (-0.45,1.85) -- (-0.15,1.65);
\node[lab, rotate=90] at (-0.68,2.7) {$T$ timesteps};
\node[lab] at (0.825,1.55) {$C$ cameras};

\coordinate (vidR) at (1.72, 2.66);
\node[minimum width=1.0cm, minimum height=1.4cm, inner sep=0pt, outer sep=0pt] (bb) at (2.65,2.66) {};
\draw[draw=black!70, fill=black!4, line width=0.3pt]
  (bb.north west) -- (bb.south west) -- ($(bb.south east)+(0,0.28)$) -- ($(bb.north east)+(0,-0.28)$) -- cycle;
\node[align=center] at (bb.center) {Image\\Backbone};
\draw[flow] (vidR) -- (bb.west);

\node[tok, fill=camA] (imgFirst) at (3.67,2.66) {};
\node[tok, fill=camB] at (3.99,2.66) {};
\node[text=black!60, inner sep=0pt] at (4.31,2.66) {$\cdots$};
\node[tok, fill=camE] at (4.63,2.66) {};
\node[tok, fill=camF] (imgLast) at (4.95,2.66) {};
\draw[flow] (bb.east) -- (imgFirst.west);
\node[lab] at (4.31,3.0) {image tokens};

\node[draw=black!70, circle, inner sep=0.5pt] (plus) at (5.6,2.66) {$+$};
\draw[flow] (imgLast.east) -- (plus.west);
\node[lab] (emb) at (5.6,3.46) {embeddings};
\draw[att] (5.6,3.22) -- (plus.north);

\node[proc] (enc) at (7.0,2.66) {Mapper \\(Self-Attention)};
\draw[flow] (plus.east) -- (enc.west);
\foreach \i in {0,1,2}{
  \pgfmathsetmacro{\col}{\scenecolors[\i]}
  \node[tok, fill=\col] at (5.66+0.32*\i, 1.75) {};
}
\node[lab] at (5.98,1.42) {map tokens (init)};
\draw[flow, rounded corners=2pt] (6.46,1.75) -| ($(enc.south)+(-0.27,0)$);

\node[tok, fill=RdYlBu4] (roFirst) at (8.45,2.66) {};
\node[tok, fill=RdYlBu3] at (8.77,2.66) {};
\node[tok, fill=RdYlBu2] (roLast) at (9.09,2.66) {};
\draw[flow] (enc.east) -- (roFirst.west);
\node[lab] at (8.77,3.0) {map tokens};

\def\discarddx{-0.15} 

\begin{scope}[transform canvas={xshift=-1mm}]
\foreach \i in {0,1}{
  \pgfmathsetmacro{\col}{\camcolors[\i]}
  \node[tok, fill=\col, fill opacity=0.2, draw=black!25, densely dashed]
    at (8.11+0.32*\i+\discarddx, 1.75) {};
}
\node[text=black!40, inner sep=0pt]
  at (8.75+\discarddx,1.75) {$\cdots$};

\foreach \i in {4,5}{
  \pgfmathsetmacro{\col}{\camcolors[\i]}
  \node[tok, fill=\col, fill opacity=0.2, draw=black!25, densely dashed]
    at (7.79+0.32*\i+\discarddx, 1.75) {};
}

\node[lab, text opacity=0.4]
  at (8.67+\discarddx,1.42) {discard image tokens};

\draw[black!45, densely dashed, -{Stealth[length=3pt]}, line width=0.45pt, rounded corners=2pt]
  ($(enc.south east)+(-0.4,0)$) |- (7.94+\discarddx,1.75);
\end{scope}

\node[dec] (decM) at (10.43,2.66) {Task-Specific\\Decoder};
\draw[flow] (roLast.east) -- (decM.west);
\foreach \i in {0,1,2}{
    \node[tok, fill=black!8] at (10.11+0.32*\i, 1.75) {};
}
\node[lab] at (10.5,1.42) {decoder queries};
\draw[flow] (10.43,1.9) -- (decM.south);

\node[draw=black!60, line width=0.3pt, fill=black!2, minimum width=1.4cm, minimum height=1.4cm, inner sep=0pt] (mapBox) at (12.31,2.66) {};
\draw[flow] (decM.east) -- (mapBox.west);

\begin{scope}[shift={(-0.44,0)}]
\draw[mapblue, line width=0.6pt, fill=mapblue!25] (12.27,2.08) -- (13.23,2.08) -- (13.23,2.26) -- (12.27,2.26) -- cycle;
\foreach \p in {(12.27,2.08),(12.59,2.08),(12.91,2.08),(13.23,2.08),(13.23,2.26),(12.91,2.26),(12.59,2.26),(12.27,2.26)}{
  \fill[mapblue] \p circle (0.55pt);
}
\draw[mapgreen, line width=0.6pt] (12.13,2.02) -- (12.22,2.22) -- (12.3,2.45) -- (12.33,2.7) -- (12.34,2.95) -- (12.33,3.18) -- (12.32,3.3);
\foreach \p in {(12.13,2.02),(12.22,2.22),(12.3,2.45),(12.33,2.7),(12.34,2.95),(12.33,3.18),(12.32,3.3)}{
  \fill[mapgreen] \p circle (0.55pt);
}
\draw[mapgreen, line width=0.6pt] (13.37,2.02) -- (13.28,2.22) -- (13.2,2.45) -- (13.17,2.7) -- (13.16,2.95) -- (13.17,3.18) -- (13.18,3.3);
\foreach \p in {(13.37,2.02),(13.28,2.22),(13.2,2.45),(13.17,2.7),(13.16,2.95),(13.17,3.18),(13.18,3.3)}{
  \fill[mapgreen] \p circle (0.55pt);
}
\draw[maporange, line width=0.6pt] (12.75,2.32) -- (12.74,2.55) -- (12.73,2.8) -- (12.74,3.05) -- (12.75,3.3);
\foreach \p in {(12.75,2.32),(12.74,2.55),(12.73,2.8),(12.74,3.05),(12.75,3.3)}{
  \fill[maporange] \p circle (0.55pt);
}
\draw[red, line width=0.5pt, rounded corners=0.4pt] (12.97-0.065,2.62-0.115) rectangle (12.97+0.065,2.62+0.115);
\draw[red, line width=0.5pt] (12.97,2.62) -- (12.97,2.62+0.115);

\draw[pink, line width=0.7pt] (12.97,2.62+0.115) .. controls (12.963,2.75) and (12.897,2.85) .. (12.88,3.02);
\draw[pink, line width=0.7pt] (12.97,2.62+0.115) -- (12.97,3.15);
\draw[pink, line width=0.7pt] (12.97,2.62+0.115) .. controls (12.98,2.75) and (13.08,2.85) .. (13.105,3.02);

\node[lab] at (12.75,3.55) {task predictions};
\end{scope}

\end{tikzpicture}%
}
\caption{\textbf{Overview of \textit{\method{}}.} A shared image backbone encodes each of the $C$ camera views over $T$ timesteps into image tokens, which are flattened and augmented with learned positional, camera, and time embeddings. Instead of projecting these tokens into a dense BEV grid, we concatenate them with $K$ learnable map tokens and jointly refine both with a self-attention mapper (Figure~\ref{fig:aggregation} (c)), without camera parameters or depth. The image tokens are then discarded, and the compact map tokens serve as the scene representation, which task-specific decoders attend to with full cross-attention. Beyond online vectorized HD map construction, this representation supports 3D object detection and downstream trajectory prediction from the predicted maps.}
\label{fig:method}
\end{figure}
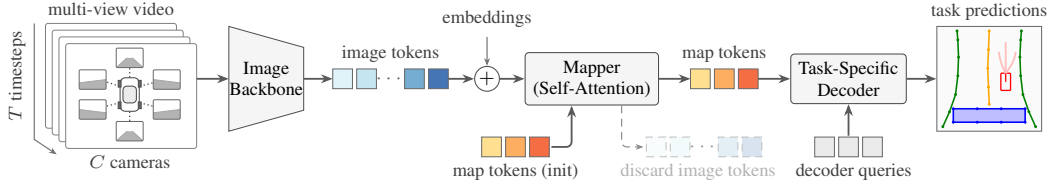


Section~\ref{subsec:task_formulation} reviews the standard BEV-based approach to online vectorized HD map construction. We then describe our modifications: flat image-token embeddings (Section~\ref{subsec:token_embeddings}), a self-attention mapper that produces compact map tokens (Section~\ref{subsec:map_mapper}), and the map decoder (Section~\ref{subsec:map_decoder}). Figure~\ref{fig:method} shows the architecture of \method{}.

\subsection{Background}
\label{subsec:task_formulation}

In this subsection, we formulate the task and review the general setup of BEV-based methods.

Given surround-view images
$\mathcal{I}=\{I_{t,c}\}_{t=1,c=1}^{T,C}$
from $C$ cameras over $T$ timesteps ($T=1$ for single-frame methods such as MapTRv2~\cite{liao2025maptrv2}), an image encoder
$f_{\mathrm{img}}$, e.g., ResNet~\cite{he2016deep} followed by a projection to $d$ channels, extracts per-view features:
\begin{equation}
    \mathbf{F}_{t,c}
    =
    f_{\mathrm{img}}(I_{t,c})
    \in
    \mathbb{R}^{H\times W\times d}.
\end{equation}

where $H$ and $W$ are the spatial dimensions of the feature map.

Existing BEV-based methods~\cite{liao2022maptr,liao2025maptrv2} then aggregate these features into a BEV grid using a projection module such as LSS~\cite{philion2020lift} or BEVFormer~\cite{li2024bevformer}:
\begin{equation}
    \mathbf{B}
    =
    f_{\mathrm{BEV}}
    \left(
        \{\mathbf{F}_{t,c}\}_{t=1,c=1}^{T,C}
    \right)
    \in
    \mathbb{R}^{N_{\mathrm{BEV}}\times d},
\end{equation}
where
$N_{\mathrm{BEV}}=H_{\mathrm{BEV}}W_{\mathrm{BEV}}$
and each token corresponds to a fixed BEV location.

A DETR-style~\cite{carion2020end} decoder then predicts $M$ map elements, each with $P$ ordered points, using queries
$\mathbf{Q}\in\mathbb{R}^{M\times P\times d}$.
In BEV-based methods, the queries retrieve information from $\mathbf{B}$ through deformable cross-attention:
\begin{equation}
    \mathbf{Q}^{(l)}
    =
    \mathbf{Q}^{(l-1)}
    +
    \operatorname{DeformAttn}
    \left(
        \mathbf{Q}^{(l-1)},
        \mathbf{B},
        \mathbf{R}^{(l-1)}
    \right),
\end{equation}
where $\mathbf{R}^{(l-1)}$ denotes the BEV sampling reference points.

Each point query is regressed to normalized coordinates
$(\hat{x},\hat{y})\in[0,1]^2$,
which are then converted to metric positions $(x,y)$ within the ego-centric perception range. Crucially, since map points are decoded from queries rather than tied to BEV cells, the dense 2D BEV grid is a design choice for the intermediate scene representation rather than an inherent requirement. This observation motivates our approach: we replace the BEV grid with a compact set of learned map tokens.

\subsection{Constructing Flat Image-Token Embeddings}
\label{subsec:token_embeddings}

We first describe how we represent the image features without a BEV grid. Instead of lifting
$\{\mathbf{F}_{t,c}\}$
into a dense BEV feature $\mathbf{B}$, we directly flatten them into
$N = TCHW$ image tokens $X \in \mathbb{R}^{N\times d}$.
Flattening, however, discards the spatial, camera, and temporal structure of the features, so we reintroduce it through learned embeddings. We encode within-image position with learnable row and column tables
$E^{row}\in\mathbb{R}^{H\times d/2}$
and
$E^{col}\in\mathbb{R}^{W\times d/2}$:
\begin{equation}
    PE^{pos}_{h,w}
    =
    [E^{row}_{h};E^{col}_{w}].
\end{equation}

We further add learnable camera and time embeddings
$E^{cam}\in\mathbb{R}^{C\times d}$
and
$E^{time}\in\mathbb{R}^{T\times d}$,
which distinguish tokens from different views and timesteps. Each token
$\mathbf{x}_{t,c,h,w}$
is thus embedded as
\begin{equation}
    \widetilde{\mathbf{x}}_{t,c,h,w}
    =
    \mathbf{x}_{t,c,h,w}
    + E^{cam}_c
    + E^{time}_t
    + PE^{pos}_{h,w}.
\end{equation}
The embedded tokens form
$\mathbf{X}^{(0)}\in\mathbb{R}^{N\times d}$,
which serves as the input to the mapper.

Unlike BEV-based methods~\cite{liao2022maptr, liao2025maptrv2}, our mapper uses no camera intrinsics, extrinsics, depth, or explicit ego-pose alignment across timesteps. Instead, camera identity, time, and within-image position are represented solely by the learned embeddings, and the mapping from image space to the map is learned entirely from data.

\subsection{Self-Attention Mapper}
\label{subsec:map_mapper}

Our mapper replaces $f_{\mathrm{BEV}}$. Following prior work using learnable latent tokens to summarize dense inputs~\cite{jin2024lvsm,zhang2025relitlrm}, we concatenate $\mathbf{X}^{(0)}$ with $K$ learnable map tokens $\mathbf{U}^{(0)} \in \mathbb{R}^{K \times d}$, where $(0)$ denotes the mapper input:
\begin{equation}
    \mathbf{Z}^{(0)} = \operatorname{Concat}\left(\mathbf{U}^{(0)}, \mathbf{X}^{(0)}\right) \in \mathbb{R}^{(K+N) \times d}.
\end{equation}

Map tokens provide a compact scene representation. We use $T=4$ (current and three past timesteps) and $K=\gamma CT$, with $\gamma=50$ by default, so $K = 1{,}200$ for nuScenes and $1{,}400$ for Argoverse~2.

Rather than updating map tokens via one-way cross-attention~\cite{jaegle2021perceiver}, we jointly refine image and map tokens with self-attention, so image tokens also exchange information across views and timesteps instead of serving as fixed keys and values. Our mapper is an $L$-layer ($L = 4$) Pre-LN Transformer~\cite{xiong2020layer}:
\begin{align}
    \widetilde{\mathbf{Z}}^{(l)}
    &=
    \mathbf{Z}^{(l-1)}
    +
    \mathrm{MSA}^{(l)}
    \bigl(
        \mathrm{LN}^{(l)}_1(\mathbf{Z}^{(l-1)})
    \bigr), \\
    \mathbf{Z}^{(l)}
    &=
    \widetilde{\mathbf{Z}}^{(l)}
    +
    \mathrm{FFN}^{(l)}
    \bigl(
        \mathrm{LN}^{(l)}_2(\widetilde{\mathbf{Z}}^{(l)})
    \bigr),
    \qquad l=1,\ldots,L,
\end{align}
where $\mathrm{MSA}$, $\mathrm{LN}$, and $\mathrm{FFN}$ denote multi-head self-attention, LayerNorm~\cite{ba2016layer}, and the feed-forward network, respectively, with
\begin{equation}
    \mathrm{FFN}(\mathbf{z})
    =
    \mathbf{W}_2
    \,\mathrm{GELU}
    \left(
        \mathbf{W}_1\mathbf{z}
        +
        \mathbf{b}_1
    \right)
    +
    \mathbf{b}_2.
\end{equation}

After the final layer, we split the output back into map and image tokens:
\begin{equation}
    \mathbf{Z}^{(L)}
    =
    \operatorname{Concat}\left(\mathbf{U}^{(L)}, \mathbf{X}^{(L)}\right).
\end{equation}
Since the map tokens serve as the compact scene representation and retaining image tokens ($N \gg K$) would be costly for decoding, we discard
$\mathbf{X}^{(L)}$
and retain only the normalized map tokens:
\begin{equation}
    \mathbf{U}
    =
    \mathrm{LN}_{\mathrm{out}}
    \bigl(
        \mathbf{U}^{(L)}
    \bigr)
    \in
    \mathbb{R}^{K\times d},
\end{equation}
where $\mathrm{LN}_{\mathrm{out}}$ is applied independently over each token's $d$ feature dimensions. 

\subsection{Map Decoder}
\label{subsec:map_decoder}

We retain the decoder architecture of MapTRv2~\cite{liao2025maptrv2}, including its $L_d = 6$ layers. We change what the decoder attends to: the compact map tokens $\mathbf{U} \in \mathbb{R}^{K \times d}$ instead of the dense BEV features $\mathbf{B} \in \mathbb{R}^{N_{\mathrm{BEV}} \times d}$. This change affects the decoder in two ways. First, since $\mathbf{U}$ has no fixed 2D coordinates, deformable attention, which samples features around BEV reference points, no longer applies. Second, since $K \ll N_{\mathrm{BEV}}$ ($1{,}200$ map tokens versus $20{,}000$ BEV tokens on nuScenes, \textbf{16.7$\times$} fewer), full cross-attention becomes feasible, letting each query attend to all map tokens.

Following MapTRv2~\cite{liao2025maptrv2}, each decoder layer first applies decoupled self-attention to the queries along the instance and point dimensions, yielding $\overline{\mathbf{Q}}^{(l)}$. We then replace deformable cross-attention with full cross-attention to the map tokens:
\begin{equation}
    \widetilde{\mathbf{Q}}^{(l)}
    =
    \overline{\mathbf{Q}}^{(l)}
    +
    \operatorname{MHA}
    \bigl(
        \overline{\mathbf{Q}}^{(l)},
        \mathbf{U},
        \mathbf{U}
    \bigr),
    \qquad l=1,\ldots,L_d,
\end{equation}
where $\mathrm{MHA}$ denotes multi-head attention with $\overline{\mathbf{Q}}^{(l)}$ as queries and $\mathbf{U}$ as keys and values.

\noindent\textbf{Prediction head.}
We retain MapTRv2's iterative coordinate refinement. Initial reference points are predicted from the decoder queries via a linear projection and sigmoid. At each layer, the regression branch predicts offsets that refine the current reference points for the next layer, while the classification branch predicts the class of each map element. Unlike MapTRv2, these reference points serve only for coordinate regression, not feature sampling. The final normalized coordinates are converted to metric positions as described in Section~\ref{subsec:task_formulation}, and predictions from all $L_d$ layers are supervised.


\noindent\textbf{Efficiency.} Our mapper and decoder use standard attention and MLP operations, so all attention layers run with FlashAttention~\cite{dao2022flashattention}, whose memory scales linearly with sequence length for both self- and cross-attention. Compute is quadratic in $N+K$, but confined to the $L=4$ mapper layers, where runtime is largely bound by memory access rather than FLOPs. Following prior work~\cite{chen2024maptracker}, we cache past-frame features, so only the current frame is encoded. Our model runs at 18.9~FPS on nuScenes (Table~\ref{tab:results_60x30}), faster than MapTRv2's dense BEV projection (14.1~FPS) and comparable to cross-attention (18.8~FPS). Breakdowns are in Appendix~\ref{app:efficiency}.


\section{Experiments}

In this section, we first introduce the dataset splits in Section~\ref{sec:datasets_splits}, followed by the evaluation metrics and baseline methods in Section~\ref{sec:metrics_baselines}. We then evaluate on online vectorized HD map construction in Section~\ref{sec:map_construction}, followed by uncertainty-aware map construction and its downstream trajectory prediction performance in Section~\ref{sec:uncertainty_traj}. The appendix provides additional qualitative comparisons (Appendix~\ref{app:quali_map}), map construction results under different data splits (Appendix~\ref{app:map}), real-time variants (Appendix~\ref{app:realtime}), 3D object detection (Appendix~\ref{app:detection}), ablation studies (Appendix~\ref{app:ablation}), data split visualizations (Appendix~\ref{app:map_split}), robustness to camera extrinsic perturbations (Appendix~\ref{app:robustness}), and implementation details (Appendix~\ref{app:implement}). \textbf{Video results are provided in the supplementary material.}


\subsection{Datasets and Splits}
\label{sec:datasets_splits}

\noindent \textbf{Datasets.} We conduct experiments on nuScenes~\cite{caesar2020nuscenes} and Argoverse 2~\cite{wilson2023argoverse}, using images from 6 and 7 cameras, respectively, together with the corresponding ground truth vectorized map elements. We use the same key frames as MapTRv2~\cite{liao2025maptrv2}. The map classes are road boundaries, lane dividers, and pedestrian crossings.

\begin{wraptable}[22]{r}{0.5\linewidth} 
\centering
\vspace{-20pt} 
\caption{\textbf{Comparison of aggregation mechanisms under a common decoder framework in the single-frame setting ($T=1$).} Self-attention achieves the best accuracy with \textbf{66.7$\times$} fewer intermediate tokens than dense BEV projection~\cite{liao2025maptrv2}, and matches cross-attention in token count and FPS, demonstrating effective token compression and interaction. Token counts (0.3k) are for nuScenes; Argoverse 2 uses 0.35k tokens for 7 cameras (same per-camera budget).}
\label{tab:mechanism}
\resizebox{\linewidth}{!}{%
\begin{tabular}{lcccc}
\toprule
Aggregation & \begin{tabular}[c]{@{}c@{}}\#Tokens\\(k)$\downarrow$\end{tabular} & FPS$\uparrow$ & \begin{tabular}[c]{@{}c@{}}nuSc.\\mAP$\uparrow$\end{tabular} & \begin{tabular}[c]{@{}c@{}}Argo2\\mAP$\uparrow$\end{tabular} \\
\midrule
\multicolumn{5}{c}{\textit{ResNet-50 backbone}} \\
\midrule
BEV Projection            & 20.0 & 14.1 & 26.7 & 53.1 \\
Ours (cross-attn)         & 0.3  & 19.1 & 25.5 & 52.6 \\
\textbf{Ours (self-attn)} & \textbf{0.3} & \textbf{19.2} & \textbf{30.1} & \textbf{58.3} \\
\midrule
\multicolumn{5}{c}{\textit{MobileNetV3-Large backbone}} \\
\midrule
BEV Projection            & 20.0 & 23.3 & 20.5 & 43.7 \\
Ours (cross-attn)         & 0.3  & 41.2 & 21.1 & 50.0 \\
\textbf{Ours (self-attn)} & \textbf{0.3} & \textbf{41.6} & \textbf{27.6} & \textbf{56.5} \\
\midrule
\multicolumn{5}{c}{\textit{ResNet-18 backbone}} \\
\midrule
BEV Projection            & 20.0 & 24.2 & 23.0 & 46.7 \\
Ours (cross-attn)         & 0.3  & 44.2 & 21.2 & 48.1 \\
\textbf{Ours (self-attn)} & \textbf{0.3} & \textbf{44.5} & \textbf{26.9} & \textbf{55.1} \\
\bottomrule
\end{tabular}
}
\end{wraptable}

\noindent \textbf{Dataset Splits.} To rigorously evaluate generalization and avoid geographic data leakage, we adopt the \textbf{Near-Extrapolation split~\cite{lilja2024localization} with a 60m$\times$30m perception range as our default benchmark} for both nuScenes and Argoverse 2, following recent protocols~\cite{lowens2025pseudomaptrainer,lyu2025maprf}. The original splits of nuScenes and Argoverse 2 suffer from severe train–val overlap (79.4\% and 45.0\%, respectively). While the StreamMapNet split~\cite{yuan2024streammapnet} reduces nuScenes overlap to 2.1\%, the Near-Extrapolation split minimizes it to 0.9\% (within 5~m) for nuScenes (see Appendix~\ref{app:map_split} for visualizations) and 0\% for Argoverse 2. For completeness, we also report improvements on the original and StreamMapNet splits in Appendix~\ref{app:map}. For 3D object detection, we follow standard practice and use the original nuScenes split, since geographic overlap is less problematic for dynamic objects than for static HD map elements that persist across traversals.

\subsection{Metrics and Baselines}
\label{sec:metrics_baselines}

\noindent \textbf{Metrics.} We report mean Average Precision (mAP) based on Chamfer distance, with thresholds of \{0.5, 1.0, 1.5\}~m. Following prior work~\cite{liao2022maptr,liao2025maptrv2}, runtime \textbf{for all methods} is reported in frames per second (FPS) on an NVIDIA GeForce RTX 3090 for a fair comparison. Best results overall are \textbf{bolded}, and the best results among prior methods are \underline{underlined}.

\noindent \textbf{Baselines.} For online vectorized HD map construction, we benchmark against methods with public codebases, including MapTR~\cite{liao2022maptr}, StreamMapNet~\cite{yuan2024streammapnet}, MapTracker~\cite{chen2024maptracker}, MapTRv2~\cite{liao2025maptrv2}, DAMap~\cite{dong2025damap}, and AugMapNet~\cite{monninger2026augmapnet}, and retrain each on the Near-Extrapolation split using its official code. VectorMapNet~\cite{liu2023vectormapnet} results are taken from~\cite{liu2023vectormapnet,lilja2024localization} because no official implementation is available. SparseMeXt~\cite{jiang2025sparsemext}, the closest prior work to ours in removing dense BEV, is excluded from our main comparisons because its code is unavailable and it reports no results on the Near-Extrapolation or StreamMapNet splits; we list its original-split result (100 epochs) in Appendix~\ref{app:map} for reference. For 3D object detection, we report baseline results from their respective papers. We use \textbf{ResNet-50~\cite{he2016deep} as the default image backbone} for both map construction and object detection.

\subsection{Online Vectorized HD Map Construction}
\label{sec:map_construction}


Table~\ref{tab:mechanism} compares the three aggregation mechanisms in Figure~\ref{fig:aggregation} under a common DETR-style decoder framework in the single-frame setting ($T=1$). With ResNet-50, our self-attention surpasses dense BEV projection by \textbf{\txtgreen{+3.4}} mAP on nuScenes and \textbf{\txtgreen{+5.2}} mAP on Argoverse 2, and cross-attention by \textbf{\txtgreen{+4.6}} mAP on nuScenes and \textbf{\txtgreen{+5.7}} mAP on Argoverse 2. Cross-attention even falls below the dense BEV baseline on both datasets (\textbf{\txtred{-1.2}} and \textbf{\txtred{-0.5}} mAP), indicating that the compression mechanism matters. All other experiments use our default setting of $T=4$, unless otherwise specified. 


Table~\ref{tab:results_60x30} shows that ours with self-attention outperforms prior methods across every category on both datasets. Ours with self-attention surpasses AugMapNet and MapTracker by \textbf{\txtgreen{+7.8}} mAP on nuScenes and MapTRv2 by \textbf{\txtgreen{+10.2}} mAP on Argoverse 2. Additionally, our self-attention variant consistently outperforms the cross-attention variant on both datasets. Ours also achieves \textbf{18.9 FPS} on nuScenes. Qualitative comparisons are shown in Figure~\ref{fig:global_main}, with additional examples in Appendix~\ref{app:quali_map}.

\begin{table}[t]
\centering
\vspace{-0.3cm}
\caption{\textbf{Quantitative comparisons of online vectorized HD map construction.} Ours with self-attention outperforms both the cross-attention variant and previous state-of-the-art methods in accuracy on nuScenes and Argoverse 2, while achieving the highest inference speed. FPS is for nuScenes.}
\vspace{1mm}
\label{tab:results_60x30}
\adjustbox{max width=\linewidth}{
\begin{tabular}{lccccccccc}
\toprule
& \multicolumn{4}{c}{nuScenes} & \multicolumn{4}{c}{Argoverse 2} & \\
\cmidrule(lr){2-5} \cmidrule(lr){6-9}
Method & AP$_\text{div}$ & AP$_\text{ped}$ & AP$_\text{bound}$ & mAP$\uparrow$ & AP$_\text{div}$ & AP$_\text{ped}$ & AP$_\text{bound}$ & mAP$\uparrow$ & FPS$\uparrow$ \\
\midrule
MapTR (\citeyear{liao2022maptr})                & 16.0 & 14.4 & 26.7 & 19.0 & 50.0 & 46.6 & 47.5 & 48.0 & \underline{15.1} \\
VectorMapNet (\citeyear{liu2023vectormapnet})   & 13.5 & 13.7 & 14.9 & 14.0 & 39.8 & 26.8 & 31.5 & 32.7 & 2.2 \\
StreamMapNet (\citeyear{yuan2024streammapnet})  & 23.0 & 25.8 & 29.5 & 26.1 & 52.7 & 49.4 & 50.0 & 50.7 & 12.5 \\
MapTracker (\citeyear{chen2024maptracker})  & 20.0 & \underline{31.3} & 31.1 & \underline{27.5} & 57.8 & 51.4 & 48.8 & 52.7 & 13.0 \\
MapTRv2 (\citeyear{liao2025maptrv2})            & 20.9 & 26.5 & 32.6 & 26.7 & \underline{58.4} & 49.7 & \underline{51.3} & \underline{53.1} & 14.1 \\
DAMap (\citeyear{dong2025damap})                & 23.5 & 17.1 & \underline{35.9} & 25.5 & 55.3 & 44.7 & 49.2 & 49.7 & 12.1 \\
AugMapNet (\citeyear{monninger2026augmapnet})   & \underline{24.7} & 28.0 & 29.9 & \underline{27.5} & 51.9 & \underline{54.0} & 50.4 & 52.1 & 11.2 \\
\midrule
\textbf{Ours} (cross-attention) & 16.8 & 32.4 & 30.0 & 26.4 & 61.9 & 54.0 & 53.5 & 56.5 & 18.8 \\ 
\textbf{Ours} \textbf{(self-attention)} & \textbf{28.3} & \textbf{41.0} & \textbf{36.7} & \textbf{35.3} & \textbf{68.2} & \textbf{63.9} & \textbf{57.9} & \textbf{63.3} & \textbf{18.9} \\
\bottomrule
\end{tabular}
}
\end{table}

\begin{figure}[t]
    \centering
     \vspace{-0.3cm}
    \includegraphics[width=\linewidth]{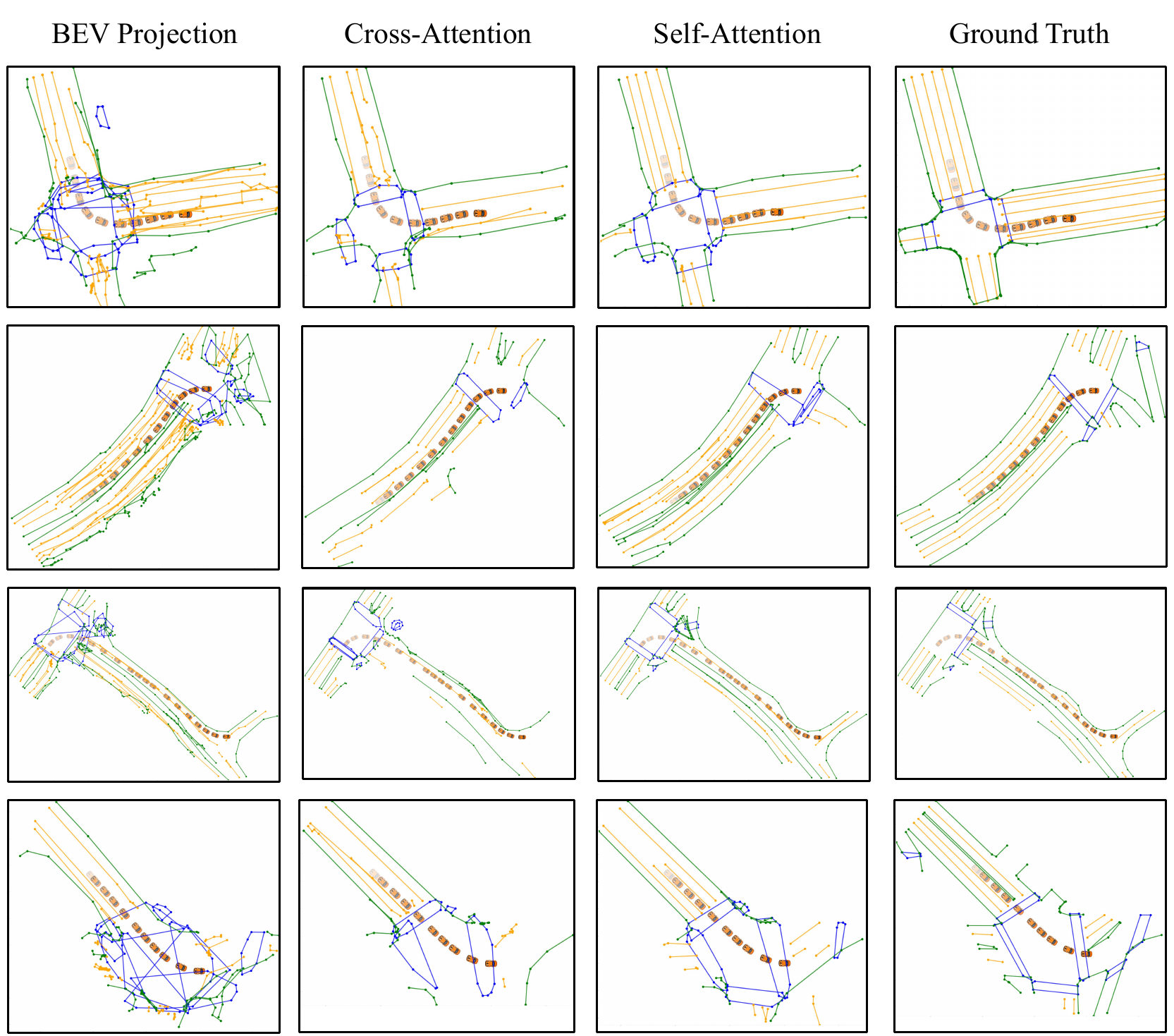}
    \vspace{-0.7cm}
    \caption{\textbf{Qualitative comparisons of online vectorized HD map construction} for Dense BEV Projection~\cite{liao2025maptrv2}, ours (cross-attention), and ours (self-attention). Self-attention produces higher-quality and more consistent vectorized map reconstructions, with fewer spurious elements, more complete map geometries, and better-preserved intersection geometry. The first three rows are from nuScenes, while the last row is from Argoverse 2. Orange, blue, and green denote lane dividers, pedestrian crossings, and road boundaries, respectively. Best viewed zoomed in. }
    \label{fig:global_main}
\end{figure}

\subsection{Uncertainty-Aware Map Construction and Trajectory Prediction}
\label{sec:uncertainty_traj}

\begin{figure}[t]
\vspace{-0.4cm}
\centering
\includegraphics[width=\linewidth]{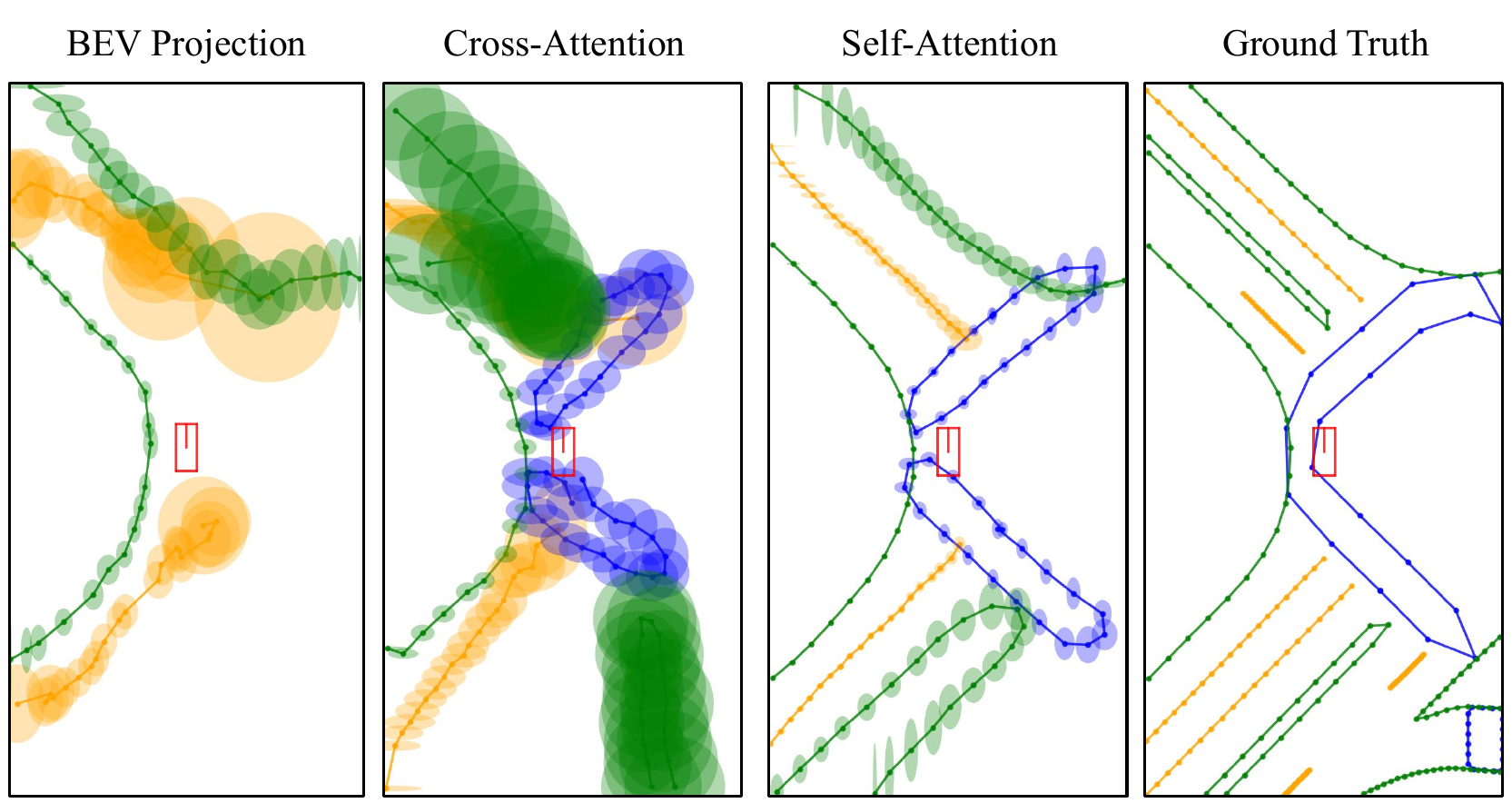}
\vspace{-0.7cm}
\caption{\textbf{Qualitative comparison of uncertainty-aware online vectorized HD map construction} for Dense BEV Projection~\cite{liao2025maptrv2}, ours (cross-attention), and ours (self-attention) on nuScenes. Uncertainty is visualized at each predicted vector point as a semi-transparent ellipse whose size reflects the predicted Laplace variance. Self-attention produces more localized predictions with substantially less diffuse uncertainty, providing cleaner uncertainty-aware maps for downstream trajectory prediction. Orange, blue, and green denote lane dividers, pedestrian crossings, and road boundaries, respectively, while the red box marks the ego vehicle. }
\label{fig:uncertainty}
\end{figure}

Table~\ref{tab:results_60x30_traj} shows that ours with self-attention surpasses MapTRv2 on uncertainty-aware map construction by \textbf{\txtgreen{+14.4}} mAP, and the cross-attention variant by \textbf{\txtgreen{+9.0}} mAP. Self-attention also improves downstream trajectory prediction across all three metrics (minADE, minFDE, and MR). Qualitative comparisons of uncertainty-aware map predictions are provided in Figure~\ref{fig:uncertainty}.

While MapUncertainty~\cite{gu2024producing} reports negligible impact on the original split, adding the uncertainty head reduces mAP for all methods on the Near-Extrapolation split~\cite{lilja2024localization}, possibly because NLL down-weights gradients for poorly predicted, high-variance samples~\cite{seitzer2022pitfalls}. Comparing Tables~\ref{tab:results_60x30} and~\ref{tab:results_60x30_traj}, our method degrades the least: \textbf{-3.0} and \textbf{-2.9} mAP for our cross- and self-attention variants, versus \textbf{\txtred{-6.8}}, \textbf{\txtred{-15.2}}, and \textbf{\txtred{-8.7}} mAP for MapTR, StreamMapNet, and MapTRv2. Although the uncertainty head slightly reduces our map construction accuracy, it enables uncertainty-aware downstream trajectory prediction~\cite{zhou2022hivt}.

\begin{table}[t]
\centering
\vspace{-0.4cm}
\caption{\textbf{Quantitative comparisons of uncertainty-aware online vectorized HD map construction and downstream trajectory prediction on nuScenes.} Following MapUncertainty~\cite{gu2024producing}, we train each map construction model with an added uncertainty head, precompute its uncertainty-aware maps, and train a separate HiVT~\cite{zhou2022hivt} on them with identical settings for trajectory prediction. We include MapTR, StreamMapNet, and MapTRv2 because MapUncertainty provides uncertainty-head implementations for them, and we add the same uncertainty head to ours as for MapTRv2. Ours with self-attention outperforms the cross-attention variant and previous methods on map construction and downstream trajectory prediction. 
}
\vspace{1mm}
\label{tab:results_60x30_traj}
\adjustbox{max width=\linewidth}{
\begin{tabular}{lccccccc}
\toprule
& \multicolumn{4}{c}{Map Construction} & \multicolumn{3}{c}{Trajectory Prediction} \\
\cmidrule(lr){2-5} \cmidrule(lr){6-8}
Method & AP$_\text{div}$ & AP$_\text{ped}$ & AP$_\text{bound}$ & mAP$\uparrow$ & minADE$\downarrow$ & minFDE$\downarrow$ & MR$\downarrow$ \\
\midrule
MapTR (\citeyear{liao2022maptr}) & 12.9 & 3.9 & 19.9 & 12.2 & \underline{0.432} & \underline{0.952} & 0.1291 \\
StreamMapNet (\citeyear{yuan2024streammapnet}) & 9.5 & \underline{11.5} & 11.8 & 10.9 & 0.440 & 0.981 & 0.1412 \\
MapTRv2 (\citeyear{liao2025maptrv2}) & \underline{16.2} & 9.7 & \underline{28.3} & \underline{18.0} & 0.435 & 0.958 & \underline{0.1274} \\
\midrule
\textbf{Ours} (cross-attention) & 17.4 & 26.3 & 26.4 & 23.4 & 0.438 & 0.979 & 0.1385 \\
\textbf{Ours} \textbf{(self-attention)} & \textbf{25.1} & \textbf{39.4} & \textbf{32.7} & \textbf{32.4} & \textbf{0.406} & \textbf{0.914} & \textbf{0.1167} \\
\bottomrule
\end{tabular}
}
\end{table}

\section{Conclusion}

We proposed \method{}, which replaces the dense 2D BEV grid with compact 1D learnable map tokens as the intermediate scene representation. Our mapper concatenates map and image tokens, applies self-attention, discards the image tokens, and retains the updated map tokens for decoding. This representation uses far fewer tokens, consumes less memory, runs faster, enables full cross-attention for global context, and avoids dependence on camera projection parameters. \method{} achieves state-of-the-art accuracy and efficiency on nuScenes and Argoverse 2, and generalizes to uncertainty-aware map construction, downstream trajectory prediction, and 3D object detection.

\noindent\textbf{Limitations.} \textbf{(i)} Replacing dense BEV features with compact 1D tokens may require additional adaptation in end-to-end frameworks, whose task decoders rely explicitly on the BEV grid structure (e.g., MapFormer and OccFormer in UniAD~\cite{hu2023planning}). \textbf{(ii)} Our fixed token budget may require tuning for other tasks or datasets, and dynamic token selection~\cite{rao2021dynamicvit,fayyaz2022adaptive,yin2022vit} is a promising direction for adapting the budget to input complexity.




\noindent \textbf{AI Disclosure.} We used generative AI to help polish writing, search for and identify related work, assist with research ideation, and edit source code. We have reviewed all AI-assisted work.

\bibliography{main}

@String(IJCV = {Int. J. Comput. Vis.})

@String(CVPR= {IEEE Conf. Comput. Vis. Pattern Recog.})

@String(ICCV= {Int. Conf. Comput. Vis.})

@String(ECCV= {Eur. Conf. Comput. Vis.})

@String(ICLR = {Int. Conf. Learn. Represent.})

@String(AAAI = {AAAI})

@String(IJCV  = {IJCV})

@String(CVPR  = {CVPR})

@String(ICCV  = {ICCV})

@String(ECCV  = {ECCV})

@String(ICLR  = {ICLR})

@inproceedings{shan2018lego,
  title={Lego-loam: Lightweight and ground-optimized lidar odometry and mapping on variable terrain},
  author={Shan, Tixiao and Englot, Brendan},
  booktitle={IROS},
  year={2018},
}

@inproceedings{zhang2014loam,
  title={LOAM: Lidar odometry and mapping in real-time.},
  author={Zhang, Ji and Singh, Sanjiv and others},
  booktitle={RSS},
  year={2014},
}

@inproceedings{zhou2022cross,
  title={Cross-view transformers for real-time map-view semantic segmentation},
  author={Zhou, Brady and Kr{\"a}henb{\"u}hl, Philipp},
  booktitle={CVPR},
  year={2022}
}

@inproceedings{carion2020end,
  title={End-to-end object detection with transformers},
  author={Carion, Nicolas and Massa, Francisco and Synnaeve, Gabriel and Usunier, Nicolas and Kirillov, Alexander and Zagoruyko, Sergey},
  booktitle={ECCV},
  year={2020},
}

@inproceedings{liu2023vectormapnet,
  title={Vectormapnet: End-to-end vectorized hd map learning},
  author={Liu, Yicheng and Yuan, Tianyuan and Wang, Yue and Wang, Yilun and Zhao, Hang},
  booktitle={ICML},
  year={2023},
}

@article{liao2022maptr,
  title={Maptr: Structured modeling and learning for online vectorized hd map construction},
  author={Liao, Bencheng and Chen, Shaoyu and Wang, Xinggang and Cheng, Tianheng and Zhang, Qian and Liu, Wenyu and Huang, Chang},
  journal={arXiv preprint arXiv:2208.14437},
  year={2022}
}

@article{liao2025maptrv2,
  title={Maptrv2: An end-to-end framework for online vectorized hd map construction},
  author={Liao, Bencheng and Chen, Shaoyu and Zhang, Yunchi and Jiang, Bo and Zhang, Qian and Liu, Wenyu and Huang, Chang and Wang, Xinggang},
  journal={IJCV},
  year={2025},
}

@inproceedings{yuan2024streammapnet,
  title={Streammapnet: Streaming mapping network for vectorized online hd map construction},
  author={Yuan, Tianyuan and Liu, Yicheng and Wang, Yue and Wang, Yilun and Zhao, Hang},
  booktitle={WACV},
  year={2024}
}

@inproceedings{chen2024maptracker,
  title={Maptracker: Tracking with strided memory fusion for consistent vector hd mapping},
  author={Chen, Jiacheng and Wu, Yuefan and Tan, Jiaqi and Ma, Hang and Furukawa, Yasutaka},
  booktitle={ECCV},
  year={2024},
}

@article{wilson2023argoverse,
  title={Argoverse 2: Next generation datasets for self-driving perception and forecasting},
  author={Wilson, Benjamin and Qi, William and Agarwal, Tanmay and Lambert, John and Singh, Jagjeet and Khandelwal, Siddhesh and Pan, Bowen and Kumar, Ratnesh and Hartnett, Andrew and Pontes, Jhony Kaesemodel and others},
  journal={arXiv preprint arXiv:2301.00493},
  year={2023}
}

@inproceedings{lilja2024localization,
  title={Localization is all you evaluate: Data leakage in online mapping datasets and how to fix it},
  author={Lilja, Adam and Fu, Junsheng and Stenborg, Erik and Hammarstrand, Lars},
  booktitle={CVPR},
  year={2024}
}

@inproceedings{dong2025damap,
  title={DAMap: Distance-aware MapNet for High Quality HD Map Construction},
  author={Dong, Jinpeng and Li, Chen and Lin, Yutong and Fu, Jingwen and Zhou, Sanping and Zheng, Nanning},
  booktitle={ICCV},
  year={2025}
}

@inproceedings{monninger2026augmapnet,
  title={Augmapnet: Improving spatial latent structure via bev grid augmentation for enhanced vectorized online hd map construction},
  author={Monninger, Thomas and Anwar, Md Zafar and Antol, Stanislaw and Staab, Steffen and Ding, Sihao},
  booktitle={WACV},
  year={2026}
}

@inproceedings{he2016deep,
  title={Deep residual learning for image recognition},
  author={He, Kaiming and Zhang, Xiangyu and Ren, Shaoqing and Sun, Jian},
  booktitle={CVPR},
  year={2016}
}

@inproceedings{caesar2020nuscenes,
  title={nuscenes: A multimodal dataset for autonomous driving},
  author={Caesar, Holger and Bankiti, Varun and Lang, Alex H and Vora, Sourabh and Liong, Venice Erin and Xu, Qiang and Krishnan, Anush and Pan, Yu and Baldan, Giancarlo and Beijbom, Oscar},
  booktitle={CVPR},
  year={2020}
}

@inproceedings{philion2020lift,
  title={Lift, splat, shoot: Encoding images from arbitrary camera rigs by implicitly unprojecting to 3d},
  author={Philion, Jonah and Fidler, Sanja},
  booktitle={ECCV},
  year={2020},
}

@article{li2024bevformer,
  title={Bevformer: learning bird's-eye-view representation from lidar-camera via spatiotemporal transformers},
  author={Li, Zhiqi and Wang, Wenhai and Li, Hongyang and Xie, Enze and Sima, Chonghao and Lu, Tong and Yu, Qiao and Dai, Jifeng},
  journal={TPAMI},
  year={2024},
}

@article{kim2024unveiling,
  title={Unveiling the hidden: Online vectorized hd map construction with clip-level token interaction and propagation},
  author={Kim, Nayeon and Seong, Hongje and Ji, Daehyun and Jang, Sujin},
  journal={NeurIPS},
  year={2024}
}

@inproceedings{jaegle2021perceiver,
  title={Perceiver: General perception with iterative attention},
  author={Jaegle, Andrew and Gimeno, Felix and Brock, Andy and Vinyals, Oriol and Zisserman, Andrew and Carreira, Joao},
  booktitle={ICML},
  year={2021},
}

@inproceedings{li2023blip,
  title={Blip-2: Bootstrapping language-image pre-training with frozen image encoders and large language models},
  author={Li, Junnan and Li, Dongxu and Savarese, Silvio and Hoi, Steven},
  booktitle={ICML},
  year={2023},
}

@inproceedings{lowens2025pseudomaptrainer,
  title={Pseudomaptrainer: Learning online mapping without hd maps},
  author={L{\"o}wens, Christian and Funke, Thorben and Xie, Jingchao and Condurache, Alexandru Paul},
  booktitle={ICCV},
  year={2025}
}

@article{lyu2025maprf,
  title={MapRF: Weakly Supervised Online HD Map Construction via NeRF-Guided Self-Training},
  author={Lyu, Hongyu and Monninger, Thomas and Perez, Julie Stephany Berrio and Shan, Mao and Ming, Zhenxing and Worrall, Stewart},
  journal={arXiv preprint arXiv:2511.19527},
  year={2025}
}

@article{yang2025towards,
  title={Towards Efficient and Effective Multi-Camera Encoding for End-to-End Driving},
  author={Yang, Jiawei and Chen, Ziyu and You, Yurong and Wang, Yan and Li, Yiming and Chen, Yuxiao and Li, Boyi and Ivanovic, Boris and Pavone, Marco and Wang, Yue},
  journal={arXiv preprint arXiv:2512.10947},
  year={2025}
}

@inproceedings{darcet2024vision,
  title={Vision transformers need registers},
  author={Darcet, Timoth{\'e}e and Oquab, Maxime and Mairal, Julien and Bojanowski, Piotr},
  booktitle={ICLR},
  year={2024}
}

@article{jin2024lvsm,
  title={Lvsm: A large view synthesis model with minimal 3d inductive bias},
  author={Jin, Haian and Jiang, Hanwen and Tan, Hao and Zhang, Kai and Bi, Sai and Zhang, Tianyuan and Luan, Fujun and Snavely, Noah and Xu, Zexiang},
  journal={arXiv preprint arXiv:2410.17242},
  year={2024}
}

@article{alayrac2022flamingo,
  title={Flamingo: a visual language model for few-shot learning},
  author={Alayrac, Jean-Baptiste and Donahue, Jeff and Luc, Pauline and Miech, Antoine and Barr, Iain and Hasson, Yana and Lenc, Karel and Mensch, Arthur and Millican, Katherine and Reynolds, Malcolm and others},
  journal={NeurIPS},
  year={2022}
}

@inproceedings{strudel2021segmenter,
  title={Segmenter: Transformer for semantic segmentation},
  author={Strudel, Robin and Garcia, Ricardo and Laptev, Ivan and Schmid, Cordelia},
  booktitle={ICCV},
  year={2021}
}

@article{locatello2020object,
  title={Object-centric learning with slot attention},
  author={Locatello, Francesco and Weissenborn, Dirk and Unterthiner, Thomas and Mahendran, Aravindh and Heigold, Georg and Uszkoreit, Jakob and Dosovitskiy, Alexey and Kipf, Thomas},
  journal={NeurIPS},
  year={2020}
}

@inproceedings{dosovitskiy2021image,
  title={An image is worth 16x16 words: Transformers for image recognition at scale},
  author={Dosovitskiy, Alexey and Beyer, Lucas and Kolesnikov, Alexander and Weissenborn, Dirk and Zhai, Xiaohua and Unterthiner, Thomas and Dehghani, Mostafa and Minderer, Matthias and Heigold, Georg and Gelly, Sylvain and Uszkoreit, Jakob and Houlsby, Neil},
  booktitle={ICLR},
  year={2021}
}

@inproceedings{jaegleperceiverio,
  title={Perceiver IO: A General Architecture for Structured Inputs \& Outputs},
  author={Jaegle, Andrew and Borgeaud, Sebastian and Alayrac, Jean-Baptiste and Doersch, Carl and Ionescu, Catalin and Ding, David and Koppula, Skanda and Zoran, Daniel and Brock, Andrew and Shelhamer, Evan and others},
  booktitle={ICLR},
  year={2022}
}

@article{burtsev2020memory,
  title={Memory transformer},
  author={Burtsev, Mikhail S and Kuratov, Yuri and Peganov, Anton and Sapunov, Grigory V},
  journal={arXiv preprint arXiv:2006.11527},
  year={2020}
}

@inproceedings{devlin2019bert,
  title={BERT: Pre-training of deep bidirectional transformers for language understanding},
  author={Devlin, Jacob and Chang, Ming-Wei and Lee, Kenton and Toutanova, Kristina},
  booktitle={NAACL-HLT},
  year={2019}
}

@article{mu2023learning,
  title={Learning to compress prompts with gist tokens},
  author={Mu, Jesse and Li, Xiang and Goodman, Noah},
  journal={NeurIPS},
  year={2023}
}

@inproceedings{chevalier2023adapting,
  title={Adapting language models to compress contexts},
  author={Chevalier, Alexis and Wettig, Alexander and Ajith, Anirudh and Chen, Danqi},
  booktitle={EMNLP},
  year={2023}
}

@inproceedings{vani2024sparo,
  title={Sparo: Selective attention for robust and compositional transformer encodings for vision},
  author={Vani, Ankit and Nguyen, Bac and Lavoie, Samuel and Krishna, Ranjay and Courville, Aaron},
  booktitle={ECCV},
  year={2024},
}

@inproceedings{gu2024producing,
  title={Producing and leveraging online map uncertainty in trajectory prediction},
  author={Gu, Xunjiang and Song, Guanyu and Gilitschenski, Igor and Pavone, Marco and Ivanovic, Boris},
  booktitle={CVPR},
  year={2024},
}

@inproceedings{zhou2022hivt,
  title={Hivt: Hierarchical vector transformer for multi-agent motion prediction},
  author={Zhou, Zikang and Ye, Luyao and Wang, Jianping and Wu, Kui and Lu, Kejie},
  booktitle={CVPR},
  year={2022},
}

@article{dao2022flashattention,
  title={Flashattention: Fast and memory-efficient exact attention with io-awareness},
  author={Dao, Tri and Fu, Dan and Ermon, Stefano and Rudra, Atri and R{\'e}, Christopher},
  journal={NeurIPS},
  year={2022}
}

@inproceedings{zhang2025relitlrm,
  title={Relitlrm: Generative relightable radiance for large reconstruction models},
  author={Zhang, Tianyuan and Kuang, Zhengfei and Jin, Haian and Xu, Zexiang and Bi, Sai and Tan, Hao and Zhang, He and Hu, Yiwei and Hasan, Milos and Freeman, William and others},
  booktitle={ICLR},
  year={2025}
}

@inproceedings{li2024bevnext,
  title={Bevnext: Reviving dense bev frameworks for 3d object detection},
  author={Li, Zhenxin and Lan, Shiyi and Alvarez, Jose M and Wu, Zuxuan},
  booktitle={CVPR},
  year={2024},
}

@inproceedings{wang2023exploring,
  title={Exploring object-centric temporal modeling for efficient multi-view 3d object detection},
  author={Wang, Shihao and Liu, Yingfei and Wang, Tiancai and Li, Ying and Zhang, Xiangyu},
  booktitle={ICCV},
  year={2023},
}

@inproceedings{liu2023sparsebev,
  title={Sparsebev: High-performance sparse 3d object detection from multi-camera videos},
  author={Liu, Haisong and Teng, Yao and Lu, Tao and Wang, Haiguang and Wang, Limin},
  booktitle={ICCV},
  year={2023},
}

@inproceedings{song2025memfusionmap,
  title={Memfusionmap: Working memory fusion for online vectorized hd map construction},
  author={Song, Jingyu and Chen, Xudong and Lu, Liupei and Li, Jie and Skinner, Katherine A},
  booktitle={WACV},
  year={2025},
}

@inproceedings{yang2025mambamap,
  title={MambaMap: Online vectorized HD map construction using state space model},
  author={Yang, Ruizi and Liu, Xiaolu and Chen, Junbo and Zhu, Jianke},
  booktitle={IROS},
  year={2025},
}

@article{seitzer2022pitfalls,
  title={On the pitfalls of heteroscedastic uncertainty estimation with probabilistic neural networks},
  author={Seitzer, Maximilian and Tavakoli, Arash and Antic, Dimitrije and Martius, Georg},
  journal={arXiv preprint arXiv:2203.09168},
  year={2022}
}

@inproceedings{li2023bevdepth,
  title={Bevdepth: Acquisition of reliable depth for multi-view 3d object detection},
  author={Li, Yinhao and Ge, Zheng and Yu, Guanyi and Yang, Jinrong and Wang, Zengran and Shi, Yukang and Sun, Jianjian and Li, Zeming},
  booktitle={AAAI},
  year={2023}
}

@article{lin2023sparse4d,
  title={Sparse4d v2: Recurrent temporal fusion with sparse model},
  author={Lin, Xuewu and Lin, Tianwei and Pei, Zixiang and Huang, Lichao and Su, Zhizhong},
  journal={arXiv preprint arXiv:2305.14018},
  year={2023}
}

@article{han2024exploring,
  title={Exploring recurrent long-term temporal fusion for multi-view 3d perception},
  author={Han, Chunrui and Yang, Jinrong and Sun, Jianjian and Ge, Zheng and Dong, Runpei and Zhou, Hongyu and Mao, Weixin and Peng, Yuang and Zhang, Xiangyu},
  journal={RA-L},
  year={2024},
}

@inproceedings{yang2023bevformer,
  title={Bevformer v2: Adapting modern image backbones to bird's-eye-view recognition via perspective supervision},
  author={Yang, Chenyu and Chen, Yuntao and Tian, Hao and Tao, Chenxin and Zhu, Xizhou and Zhang, Zhaoxiang and Huang, Gao and Li, Hongyang and Qiao, Yu and Lu, Lewei and others},
  booktitle={CVPR},
  year={2023},
}

@article{loshchilov2017decoupled,
  title={Decoupled weight decay regularization},
  author={Loshchilov, Ilya and Hutter, Frank},
  journal={arXiv preprint arXiv:1711.05101},
  year={2017}
}

@inproceedings{howard2019searching,
  title={Searching for mobilenetv3},
  author={Howard, Andrew and Sandler, Mark and Chu, Grace and Chen, Liang-Chieh and Chen, Bo and Tan, Mingxing and Wang, Weijun and Zhu, Yukun and Pang, Ruoming and Vasudevan, Vijay and others},
  booktitle={ICCV},
  year={2019}
}

@article{zhou2021deepvit,
  title={Deepvit: Towards deeper vision transformer},
  author={Zhou, Daquan and Kang, Bingyi and Jin, Xiaojie and Yang, Linjie and Lian, Xiaochen and Jiang, Zihang and Hou, Qibin and Feng, Jiashi},
  journal={arXiv preprint arXiv:2103.11886},
  year={2021}
}

@article{wang2025efficient,
  title={Efficient and robust multi-camera 3D object detection in bird-eye-view},
  author={Wang, Yuanlong and Jiang, Hengtao and Chen, Guanying and Zhang, Tong and Zhou, Jiaqing and Qing, Zezheng and Wang, Chunyan and Zhao, Wanzhong},
  journal={IVC},
  year={2025},
}

@article{rao2021dynamicvit,
  title={Dynamicvit: Efficient vision transformers with dynamic token sparsification},
  author={Rao, Yongming and Zhao, Wenliang and Liu, Benlin and Lu, Jiwen and Zhou, Jie and Hsieh, Cho-Jui},
  journal={NeurIPS},
  year={2021}
}

@inproceedings{fayyaz2022adaptive,
  title={Adaptive token sampling for efficient vision transformers},
  author={Fayyaz, Mohsen and Koohpayegani, Soroush Abbasi and Jafari, Farnoush Rezaei and Sengupta, Sunando and Joze, Hamid Reza Vaezi and Sommerlade, Eric and Pirsiavash, Hamed and Gall, J{\"u}rgen},
  booktitle={ECCV},
  year={2022},
}

@inproceedings{yin2022vit,
  title={A-vit: Adaptive tokens for efficient vision transformer},
  author={Yin, Hongxu and Vahdat, Arash and Alvarez, Jose M and Mallya, Arun and Kautz, Jan and Molchanov, Pavlo},
  booktitle={CVPR},
  year={2022},
}

@inproceedings{hu2023planning,
  title={Planning-oriented autonomous driving},
  author={Hu, Yihan and Yang, Jiazhi and Chen, Li and Li, Keyu and Sima, Chonghao and Zhu, Xizhou and Chai, Siqi and Du, Senyao and Lin, Tianwei and Wang, Wenhai and others},
  booktitle={CVPR},
  year={2023},
}

@inproceedings{su2025freqpde,
  title={Freqpde: Rethinking positional depth embedding for multi-view 3d object detection transformers},
  author={Su, Haisheng and Zhang, Junjie and Song, Feixiang and Zhou, Sanping and Wu, Wei and Yan, Junchi and Zheng, Nanning},
  booktitle={ICCV},
  year={2025},
}

@inproceedings{fan2026stur3d,
  title={STUR3D: Spatio-Temporal Unified Representation Learning for 3D Object Detection},
  author={Fan, Huijie and Huang, Pengrui and Wang, Qiang and Fan, Baojie and Dong, Jiahua and Qu, Liangqiong},
  booktitle={CVPR},
  year={2026}
}

@article{zhu2020deformable,
  title={Deformable detr: Deformable transformers for end-to-end object detection},
  author={Zhu, Xizhou and Su, Weijie and Lu, Lewei and Li, Bin and Wang, Xiaogang and Dai, Jifeng},
  journal={arXiv preprint arXiv:2010.04159},
  year={2020}
}

@inproceedings{jiang2025sparsemext,
  title={Sparsemext: unlocking the potential of sparse representations for HD map construction},
  author={Jiang, Anqing and Chai, Jinhao and Gao, Yu and Wang, Yiru and Heng, Yuwen and Sun, Zhigang and Sun, Hao and Zhao, Zezhong and Sun, Li and Zhou, Jian and others},
  booktitle={IROS},
  year={2025},
}

@article{yu2023flashocc,
  title={Flashocc: Fast and memory-efficient occupancy prediction via channel-to-height plugin},
  author={Yu, Zichen and Shu, Changyong and Deng, Jiajun and Lu, Kangjie and Liu, Zongdai and Yu, Jiangyong and Yang, Dawei and Li, Hui and Chen, Yan},
  journal={arXiv preprint arXiv:2311.12058},
  year={2023}
}

@inproceedings{hou2024fastocc,
  title={Fastocc: Accelerating 3d occupancy prediction by fusing the 2d bird’s-eye view and perspective view},
  author={Hou, Jiawei and Li, Xiaoyan and Guan, Wenhao and Zhang, Gang and Feng, Di and Du, Yuheng and Xue, Xiangyang and Pu, Jian},
  booktitle={ICRA},
  year={2024},
}

@inproceedings{xiong2020layer,
  title={On layer normalization in the transformer architecture},
  author={Xiong, Ruibin and Yang, Yunchang and He, Di and Zheng, Kai and Zheng, Shuxin and Xing, Chen and Zhang, Huishuai and Lan, Yanyan and Wang, Liwei and Liu, Tieyan},
  booktitle={ICML},
  year={2020},
}

@article{ba2016layer,
  title={Layer normalization},
  author={Ba, Jimmy Lei and Kiros, Jamie Ryan and Hinton, Geoffrey E},
  journal={arXiv preprint arXiv:1607.06450},
  year={2016}
}

@article{zhanabatyrova2024structure,
  title={Structure from motion-based mapping for autonomous driving: practice and experience},
  author={Zhanabatyrova, Aziza and Leite, Clayton Souza and Xiao, Yu},
  journal={TIOT},
  year={2024},
}

@inproceedings{lin2014microsoft,
  title={Microsoft coco: Common objects in context},
  author={Lin, Tsung-Yi and Maire, Michael and Belongie, Serge and Hays, James and Perona, Pietro and Ramanan, Deva and Doll{\'a}r, Piotr and Zitnick, C Lawrence},
  booktitle={ECCV},
  year={2014},
}

@inproceedings{deng2009imagenet,
  title={Imagenet: A large-scale hierarchical image database},
  author={Deng, Jia and Dong, Wei and Socher, Richard and Li, Li-Jia and Li, Kai and Fei-Fei, Li},
  booktitle={CVPR},
  year={2009},
}
\bibliographystyle{template/iclr2026_conference}

\newpage
\appendix
\clearpage

\section{Additional Qualitative Results}

\subsection{Online Vectorized HD Map Construction}
\label{app:quali_map}

\begin{figure}[ht]
    \centering
    \includegraphics[width=\linewidth]{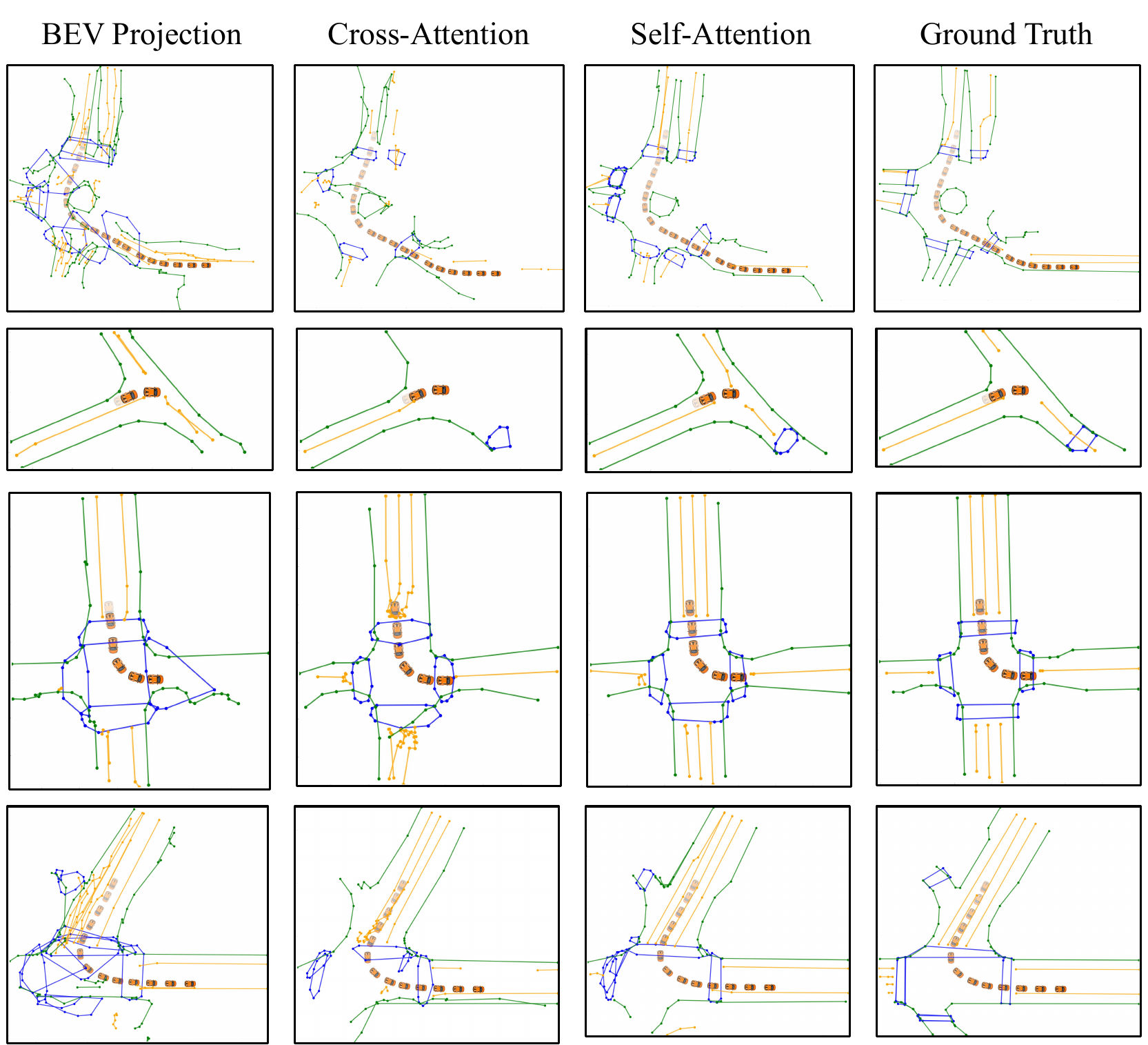}\\
    \caption{\textbf{Qualitative comparisons of online vectorized HD map construction} for Dense BEV Projection~\cite{liao2025maptrv2}, ours (cross-attention), and ours (self-attention). Self-attention produces higher-quality and more consistent vectorized map reconstructions, with fewer spurious elements, more complete map geometries, and better-preserved intersection geometry. The top two examples are from nuScenes, while the bottom two are from Argoverse 2. Orange, blue, and green denote lane dividers, pedestrian crossings, and road boundaries, respectively. Best viewed zoomed in. }
    \label{fig:global}
\end{figure}


\clearpage
\section{Additional Quantitative Results}
\label{app:quant}

\subsection{Map Construction under Different Splits}
\label{app:map}

Table~\ref{tab:results_smn_split} shows that, under the StreamMapNet split, ours surpasses MapTracker by \textbf{\txtgreen{+5.5}} mAP on nuScenes and \textbf{\txtgreen{+2.8}} mAP on Argoverse 2 at 60m$\times$30m, and by \textbf{\txtgreen{+5.9}} and \textbf{\txtgreen{+2.4}} mAP, respectively, at 100m$\times$50m. 

\begin{table}[ht]
\centering
\caption{\textbf{Quantitative comparisons of online vectorized HD map construction on the StreamMapNet split~\cite{yuan2024streammapnet}.} VectorMapNet, MapTracker, and AugMapNet results are taken from their original papers, and MapTR and StreamMapNet results are from~\citet{yuan2024streammapnet}. AugMapNet reports no Argoverse~2 results, and MapTRv2 and DAMap report none on this split.}
\vspace{1mm}
\label{tab:results_smn_split}
\adjustbox{max width=\linewidth}{
\begin{tabular}{lcccccccc}
\toprule
& \multicolumn{4}{c}{nuScenes} & \multicolumn{4}{c}{Argoverse 2} \\
\cmidrule(lr){2-5} \cmidrule(lr){6-9}
Method & AP$_\text{div}$ & AP$_\text{ped}$ & AP$_\text{bound}$ & mAP$\uparrow$ & AP$_\text{div}$ & AP$_\text{ped}$ & AP$_\text{bound}$ & mAP$\uparrow$ \\
\midrule
\multicolumn{9}{c}{\textit{60m $\times$ 30m perception range}} \\
\midrule
MapTR~\cite{liao2022maptr}               & 20.7 & 6.4  & 35.5 & 20.9 & 50.4 & 48.1 & 55.0 & 51.1 \\
VectorMapNet~\cite{liu2023vectormapnet}  & 17.0 & 15.8 & 21.2 & 18.0 & 34.9 & 35.6 & 37.8 & 36.1 \\
StreamMapNet~\cite{yuan2024streammapnet} & 30.1 & 29.6 & 41.9 & 33.9 & 55.9 & 56.9 & 61.4 & 58.1 \\
MapTracker~\cite{chen2024maptracker}     & 30.0 & \underline{45.9} & 45.1 & \underline{40.3} & \underline{75.1} & \underline{70.0} & \underline{68.9} & \underline{71.3} \\
AugMapNet~\cite{monninger2026augmapnet}  & \underline{30.3} & 39.4 & \underline{45.3} & 38.3 & \textemdash & \textemdash & \textemdash & \textemdash \\
\textbf{Ours (self-attention)} & \textbf{36.1} & \textbf{49.6} & \textbf{51.7} & \textbf{45.8} & \textbf{79.2} & \textbf{72.8} & \textbf{70.2} & \textbf{74.1} \\
\midrule
\multicolumn{9}{c}{\textit{100m $\times$ 50m perception range}} \\
\midrule
MapTR~\cite{liao2022maptr}               & 16.0 & 8.3  & 20.0 & 14.8 & 36.3 & 46.3 & 38.0 & 40.2 \\
VectorMapNet~\cite{liu2023vectormapnet}  & 8.1  & 12.0 & 6.3  & 8.8  & 20.6 & 32.4 & 24.3 & 25.7 \\
StreamMapNet~\cite{yuan2024streammapnet} & 19.6 & 24.8 & 24.7 & 23.0 & 44.4 & 60.5 & 48.6 & 51.2 \\
MapTracker~\cite{chen2024maptracker}     & \underline{24.3} & \underline{45.9} & \underline{38.4} & \underline{36.2} & \underline{64.6} & \underline{71.2} & \underline{58.5} & \underline{64.8} \\
AugMapNet~\cite{monninger2026augmapnet}  & 22.8 & 35.5 & 28.4 & 28.9 & \textemdash    & \textemdash    & \textemdash    & \textemdash    \\
\textbf{Ours (self-attention)} & \textbf{33.9} & \textbf{51.9} & \textbf{40.4} & \textbf{42.1} & \textbf{67.1} & \textbf{73.9} & \textbf{60.7} & \textbf{67.2} \\
\bottomrule
\end{tabular}
}
\end{table}

Table~\ref{tab:results_original_split} shows that, under the original split, ours surpasses MapTracker by \textbf{\txtgreen{+1.6}} mAP on nuScenes and \textbf{\txtgreen{+1.1}} mAP on Argoverse 2. As discussed in Section~\ref{sec:datasets_splits} in the main paper, substantial train--val geographic overlap allows models to encounter the same static map regions during training and evaluation. As a result, performance reflects memorization in addition to geographic generalization, which helps explain the smaller gains.

\begin{table}[ht]
\centering
\caption{\textbf{Quantitative comparisons of online vectorized HD map construction on the original split.} MapTR's Argoverse 2 result is taken from MapUnveiler~\cite{kim2024unveiling}. SparseMeXt uses a different 100-epoch training schedule and reports only nuScenes results; the remaining results are taken from each method's original paper. }
\vspace{1mm}
\label{tab:results_original_split}
\adjustbox{max width=\linewidth}{
\begin{tabular}{lcccccccc}
\toprule
& \multicolumn{4}{c}{nuScenes} & \multicolumn{4}{c}{Argoverse 2} \\
\cmidrule(lr){2-5} \cmidrule(lr){6-9}
Method & AP$_\text{div}$ & AP$_\text{ped}$ & AP$_\text{bound}$ & mAP$\uparrow$ & AP$_\text{div}$ & AP$_\text{ped}$ & AP$_\text{bound}$ & mAP$\uparrow$ \\
\midrule
MapTR~\cite{liao2022maptr}               & 51.5 & 46.3 & 53.1 & 50.3 & 58.1 & 54.7 & 56.7 & 56.5 \\
VectorMapNet~\cite{liu2023vectormapnet}  & 47.3 & 36.1 & 39.3 & 40.9 & 36.1 & 38.3 & 39.2 & 37.9 \\
StreamMapNet~\cite{yuan2024streammapnet} & 66.3 & 61.9 & 62.1 & 63.4 & 74.2 & 70.5 & 66.1 & 70.3 \\
MapTracker~\cite{chen2024maptracker}     & \underline{74.1} & \textbf{80.0} & \underline{74.1} & \underline{76.1} & \underline{80.0} & \underline{77.0} & \underline{73.7} & \underline{76.9} \\
MapTRv2~\cite{liao2025maptrv2}           & 62.4 & 59.8 & 62.4 & 61.5 & 72.1 & 62.9 & 67.1 & 67.4 \\
DAMap~\cite{dong2025damap}               & 64.7 & 58.5 & 65.1 & 62.8 & 73.2 & 64.7 & 70.2 & 69.4 \\
SparseMeXt~\cite{jiang2025sparsemext} & 67.0 & 62.6 & 66.1 & 65.2 & -- & -- & -- & -- \\
AugMapNet~\cite{monninger2026augmapnet}  & 65.4 & 61.9 & 63.6 & 63.6 & 57.4 & 57.4 & 61.6 & 58.8 \\
\textbf{Ours (self-attention)} & \textbf{77.2} & 76.5 & \textbf{79.5} & \textbf{77.7} & \textbf{80.2} & \textbf{78.5} & \textbf{75.3} & \textbf{78.0} \\
\bottomrule
\end{tabular}
}
\end{table}

\clearpage
\subsection{Real-Time Map Construction and Trajectory Prediction}
\label{app:realtime}

Table~\ref{tab:results_60x30_RT} shows that, with a MobileNetV3-Large~\cite{howard2019searching} backbone, ours with self-attention surpasses MapTRv2~\cite{liao2025maptrv2} by \textbf{\txtgreen{+10.1}} mAP on nuScenes and \textbf{\txtgreen{+16.2}} mAP on Argoverse 2, and the cross-attention variant by \textbf{\txtgreen{+6.5}} and \textbf{\txtgreen{+5.5}} mAP, respectively. With a ResNet-18 backbone, ours surpasses MapTRv2 by \textbf{\txtgreen{+7.3}} mAP on nuScenes and \textbf{\txtgreen{+11.5}} mAP on Argoverse 2, and the cross-attention variant by \textbf{\txtgreen{+6.3}} and \textbf{\txtgreen{+7.1}} mAP, respectively. Ours also achieves the highest inference speed with both backbones, reaching \textbf{40.2 FPS} with MobileNetV3-Large and \textbf{43.0 FPS} with ResNet-18.

\begin{table}[ht]
\centering
\caption{\textbf{Quantitative comparisons of online vectorized HD map construction using real-time image backbones.} Ours with self-attention surpasses both the cross-attention variant and MapTRv2 on nuScenes and Argoverse 2, while also achieving the highest inference speed (FPS) on nuScenes.}
\vspace{1mm}
\label{tab:results_60x30_RT}
\adjustbox{max width=\linewidth}{
\begin{tabular}{lccccccccc}
\toprule
& \multicolumn{4}{c}{nuScenes} & \multicolumn{4}{c}{Argoverse 2} & \\
\cmidrule(lr){2-5} \cmidrule(lr){6-9}
Method & AP$_\text{div}$ & AP$_\text{ped}$ & AP$_\text{bound}$ & mAP$\uparrow$ & AP$_\text{div}$ & AP$_\text{ped}$ & AP$_\text{bound}$ & mAP$\uparrow$ & FPS$\uparrow$ \\
\midrule
\multicolumn{10}{c}{\textit{MobileNetV3-Large backbone}} \\
\midrule
MapTRv2~\cite{liao2025maptrv2} & 19.2 & 13.0 & 29.2 & 20.5 & 50.7 & 37.2 & 43.2 & 43.7 & 23.3 \\
\textbf{Ours} (cross-attention) & 14.6 & 30.1 & 27.5 & 24.1 & 57.5 & 54.2  & 51.4  & 54.4 & 39.8 \\
\textbf{Ours}  \textbf{(self-attention)} & \textbf{25.4} & \textbf{34.9} & \textbf{31.6} & \textbf{30.6} & \textbf{65.3} & \textbf{61.1} & \textbf{53.2} & \textbf{59.9} & \textbf{40.2} \\
\midrule

\multicolumn{10}{c}{\textit{ResNet-18 backbone}} \\
\midrule
MapTRv2~\cite{liao2025maptrv2} & 20.3 & 15.7 & 33.0 & 23.0 & 53.4 & 41.7 & 45.1 & 46.7 & 24.2 \\
\textbf{Ours} (cross-attention) & 15.9 & 28.1 & 28.1 & 24.0 & 56.4 & 48.6 & 48.3 & 51.1 & 42.5 \\
\textbf{Ours} \textbf{(self-attention)} & \textbf{23.0} & \textbf{32.5} & \textbf{35.3} & \textbf{30.3} & \textbf{64.5} & \textbf{57.2} & \textbf{52.8} & \textbf{58.2} & \textbf{43.0} \\

\bottomrule
\end{tabular}
}
\end{table}

Table~\ref{tab:results_60x30_traj_RT} shows that, for uncertainty-aware map construction, ours with self-attention surpasses MapTRv2 by \textbf{\txtgreen{+10.7}} mAP with a MobileNetV3-Large backbone and \textbf{\txtgreen{+10.6}} mAP with a ResNet-18 backbone, and the cross-attention variant by \textbf{\txtgreen{+6.3}} and \textbf{\txtgreen{+6.1}} mAP, respectively. For downstream trajectory prediction, ours with self-attention achieves the best minADE, minFDE, and MR across both backbones.

\begin{table}[ht]
\centering
\caption{\textbf{Quantitative comparisons of uncertainty-aware online vectorized HD map construction and downstream trajectory prediction on nuScenes using real-time image backbones.} Ours with self-attention surpasses both the cross-attention variant and MapTRv2 on both map construction and downstream trajectory prediction across both image backbones.
}
\vspace{1mm}
\label{tab:results_60x30_traj_RT}
\adjustbox{max width=\linewidth}{
\begin{tabular}{lccccccc}
\toprule
& \multicolumn{4}{c}{Map Construction} & \multicolumn{3}{c}{Trajectory Prediction} \\
\cmidrule(lr){2-5} \cmidrule(lr){6-8}
Method & AP$_\text{div}$ & AP$_\text{ped}$ & AP$_\text{bound}$ & mAP$\uparrow$ & minADE$\downarrow$ & minFDE$\downarrow$ & MR$\downarrow$ \\
\midrule
\multicolumn{8}{c}{\textit{MobileNetV3-Large backbone}} \\
\midrule
MapTRv2~\cite{liao2025maptrv2} & 15.2 & 9.5 & 25.7 & 16.8 & 0.438 & 0.969 & 0.1330 \\
\textbf{Ours} (cross-attention) & 16.1 & 24.1 & 23.5 & 21.2 & 0.447 & 0.987 & 0.1351 \\
\textbf{Ours} \textbf{(self-attention)} & \textbf{24.1} & \textbf{30.0} & \textbf{28.4} & \textbf{27.5} & \textbf{0.413} & \textbf{0.933} & \textbf{0.1271} \\
\midrule
\multicolumn{8}{c}{\textit{ResNet-18 backbone}} \\
\midrule
MapTRv2~\cite{liao2025maptrv2} & 16.0 & 8.0 & 27.1 & 17.0 & 0.440 & 0.965 & 0.1299 \\
\textbf{Ours} (cross-attention) & 15.0 & 26.9 & 22.7 & 21.5 & 0.444 & 0.987 & 0.1350 \\
\textbf{Ours} \textbf{(self-attention)} & \textbf{20.6} & \textbf{31.9} & \textbf{30.2} & \textbf{27.6} & \textbf{0.420} & \textbf{0.942} & \textbf{0.1274} \\
\bottomrule
\end{tabular}
}
\end{table}

\clearpage
\subsection{3D Object Detection}
\label{app:detection}

Table~\ref{tab:bev_det} shows that replacing the BEV projection modules of BEVFormer v2~\cite{yang2023bevformer} with our self-attention mapper improves \textbf{\txtgreen{+2.6}} NDS and \textbf{\txtgreen{+3.7}} mAP, while increasing inference speed from 8.1 to \textbf{26.9 FPS} ($3.3\times$ faster). Even compared with the recent state-of-the-art STUR3D~\cite{fan2026stur3d}, ours improves \textbf{\txtgreen{+0.5}} NDS and \textbf{\txtgreen{+1.2}} mAP, while achieving lower errors across all five nuScenes metrics: mATE, mASE, mAOE, mAVE, and mAAE. Ours with self-attention also substantially outperforms the cross-attention variant in accuracy.

\begin{table}[ht]
\centering
\caption{\textbf{Quantitative comparisons of 3D object detection on the nuScenes original split with a ResNet-50 backbone.} Ours with self-attention surpasses both the cross-attention variant and state-of-the-art methods in accuracy, and achieves the highest inference speed (FPS) among methods that report it. Following common 3D object detection protocol, NDS and mAP are reported as fractions in $[0,1]$, whereas improvements in the text are in percentage points. All listed results are without perspective-view pre-training on nuImages. All methods use 256$\times$704 inputs, except BEVFormer v2, which uses 640$\times$1600 inputs and future frames as in its original paper. FPS of BEVFormer v2 and ours are measured on an NVIDIA RTX 3090; other FPS values are as reported in the original papers on RTX 3090. ``--'' indicates FPS not reported for this setting.}
\vspace{1mm}
\label{tab:bev_det}
\resizebox{\linewidth}{!}{%
\begin{tabular}{lcccccccc}
\toprule
Method & NDS$\uparrow$ & mAP$\uparrow$ & mATE$\downarrow$ & mASE$\downarrow$ & mAOE$\downarrow$ & mAVE$\downarrow$ & mAAE$\downarrow$ & FPS$\uparrow$ \\
\midrule
BEVDepth (\citeyear{li2023bevdepth}) & 0.475 & 0.351 & 0.639 & 0.267 & 0.479 & 0.428 & 0.198 & 15.7 \\
Sparse4Dv2 (\citeyear{lin2023sparse4d}) & 0.539 & 0.439 & 0.598 & 0.270 & 0.475 & 0.282 & \underline{0.179} & 20.3 \\
StreamPETR (\citeyear{wang2023exploring}) & 0.540 & 0.432 & 0.581 & 0.272 & 0.413 & 0.295 & 0.195 & \underline{26.7} \\
SparseBEV (\citeyear{liu2023sparsebev}) & 0.545 & 0.432 & 0.606 & 0.274 & \underline{0.387} & \underline{0.251} & 0.186 & 23.5 \\
BEVFormer v2 (\citeyear{yang2023bevformer}) & 0.529 & 0.423 & 0.618 & 0.273 & 0.413 & 0.333 & 0.181 & 8.1 \\
VideoBEV (\citeyear{han2024exploring}) & 0.535 & 0.422 & 0.564 & 0.276 & 0.440 & 0.286 & 0.198 & \text{--} \\
BEVNeXt (\citeyear{li2024bevnext}) & 0.548 & 0.437 & \underline{0.550} & \underline{0.265} & 0.427 & 0.260 & 0.208 & \text{--} \\
FreqPDE (\citeyear{su2025freqpde}) & 0.543 & 0.435 & 0.577 & 0.270 & 0.442 & 0.257 & 0.199 & \text{--} \\
STUR3D (\citeyear{fan2026stur3d}) & \underline{0.550} & \underline{0.448} & 0.558 & 0.275 & 0.431 & 0.257 & 0.208 & \text{--} \\
\midrule
\textbf{Ours} (cross-attention) & 0.463 & 0.390 & 0.710 & 0.271 & 0.403 & 0.355 & 0.185 & 26.7 \\
\textbf{Ours (self-attention)} & \textbf{0.555} & \textbf{0.460} & \textbf{0.518} & \textbf{0.262} & \textbf{0.366} & \textbf{0.236} & \textbf{0.170} & \textbf{26.9} \\
\bottomrule
\end{tabular}
}
\end{table}


\clearpage
\subsection{Efficiency Comparisons}
\label{app:efficiency}

\begin{table}[ht]
\centering
\caption{\textbf{Efficiency comparison with state-of-the-art methods on nuScenes.} Our \method{} uses substantially fewer intermediate tokens and achieves higher inference speed than prior methods. While existing approaches use 5k--20k tokens, ours requires only 1.2k tokens. With self-attention, our method achieves the highest FPS within each backbone: 18.9 FPS with ResNet-50, 40.2 FPS with MobileNetV3-Large, and 43.0 FPS with ResNet-18.}
\vspace{1mm}
\label{tab:efficiency}
\adjustbox{max width=\linewidth}{
\begin{tabular}{llllrr}
\toprule
Method & Backbone & Projector & Decoder & \# Tokens (k)$\downarrow$ & FPS$\uparrow$ \\
\midrule
VectorMapNet~\cite{liu2023vectormapnet} & ResNet-50 & IPM        & ART         & 20.0 & 2.2  \\
MapTR~\cite{liao2022maptr}        & ResNet-50 & GKT        & Deformable DETR        & 20.0 & 15.1 \\
StreamMapNet~\cite{yuan2024streammapnet} & ResNet-50 & BEVFormer  & Deformable DETR        & 5.0  & 12.5 \\
MapTracker~\cite{chen2024maptracker} & ResNet-50 & BEVFormer & Deformable DETR & 5.0 & 13.0 \\
MapTRv2~\cite{liao2025maptrv2}      & ResNet-50 & LSS        & Deformable DETR        & 20.0 & 14.1 \\
DAMap~\cite{dong2025damap}        & ResNet-50 & LSS        & Deformable DETR        & 20.0 & 12.1 \\
AugMapNet~\cite{monninger2026augmapnet}    & ResNet-50 & BEVFormer  & Deformable DETR        & 5.0  & 11.2 \\
\textbf{Ours}  (cross-attention)      & ResNet-50 & Cross-Attention & DETR & \textbf{1.2}    & 18.8  \\
\textbf{Ours} \textbf{(self-attention)}         & ResNet-50 & Self-Attention & DETR & \textbf{1.2}    & \textbf{18.9} \\
\midrule
MapTRv2~\cite{liao2025maptrv2} & MobileNetV3-Large & LSS & Deformable DETR & 20.0 & 23.3 \\
\textbf{Ours}  (cross-attention)       & MobileNetV3-Large & Cross-Attention & DETR & \textbf{1.2}  & 39.8 \\
\textbf{Ours}  \textbf{(self-attention)}      & MobileNetV3-Large & Self-Attention & DETR & \textbf{1.2}    & \textbf{40.2} \\
\midrule
MapTRv2~\cite{liao2025maptrv2} & ResNet-18 & LSS & Deformable DETR & 20.0 & 24.2 \\
\textbf{Ours}  (cross-attention)       & ResNet-18 & Cross-Attention & DETR & \textbf{1.2}  & 42.5 \\
\textbf{Ours}  \textbf{(self-attention)}      & ResNet-18 & Self-Attention & DETR & \textbf{1.2}    & \textbf{43.0} \\
\bottomrule
\end{tabular}
}
\end{table}

\begin{table}[ht]
\centering
\caption{\textbf{Component-wise runtime (ms) and allocated memory (MiB) on nuScenes.} We compare MapTRv2 with ours across different backbones, intermediate token counts, and numbers of mapper layers. Our design substantially reduces the runtime and memory of both the mapper and decoder, while leaving the backbone cost unchanged. The cross-attention variant has comparable cost to our self-attention design. MapTRv2 uses LSS rather than transformer layers, so its Mapper-layer count is denoted by ``--''. Components are rounded individually.}
\vspace{1mm}
\label{tab:component_breakdown}
\adjustbox{max width=\linewidth}{
\begin{tabular}{lrcrrrrrrrr}
\toprule
& & & \multicolumn{4}{c}{Runtime (ms)$\downarrow$} & \multicolumn{4}{c}{Memory (MiB)$\downarrow$} \\
\cmidrule(lr){4-7} \cmidrule(lr){8-11}
Method & \begin{tabular}[c]{@{}r@{}}\#Tokens\\(k)\end{tabular} & \begin{tabular}[c]{@{}r@{}}\#Mapper\\Layers\end{tabular}
  & Backbone & Mapper & Decoder & Total
  & Backbone & Mapper & Decoder & Total \\
\midrule
\multicolumn{11}{c}{\textit{ResNet-50 backbone}} \\
\midrule
MapTRv2~\cite{liao2025maptrv2} & 20.0 & --   & 42.4 & 11.6 & 16.9 & 70.9 & 590.9 & 363.2 & 126.5 & 1080.6 \\
Ours (cross-attention)  & 1.2  & 4    & 42.4 & 1.9  & 8.9  & 53.2 & 590.9 & 22.8  & 28.7  & 642.4  \\
Ours (self-attention) & 0.9 & 4    & 42.4 & 1.6  & 8.7  & 52.7 & 590.9 & 32.7  & 28.6  & 652.2  \\
Ours (self-attention) & 1.5 & 4    & 42.4 & 1.6  & 8.9  & 52.9 & 590.9 & 33.5  & 28.7  & 653.1  \\
Ours (self-attention) & 1.2 & 3    & 42.4 & 1.3  & 8.9  & 52.5 & 590.9 & 31.6  & 28.7  & 651.2  \\
Ours (self-attention) & 1.2 & 5    & 42.4 & 1.9  & 8.9  & 53.2 & 590.9 & 34.6  & 28.7  & 654.2  \\
Ours (self-attention) & 1.2 & 4 & 42.4 & 1.7 & 8.9 & 52.9 & 590.9 & 33.1 & 28.7 & 652.7 \\
\midrule
\multicolumn{11}{c}{\textit{MobileNetV3-Large backbone}} \\
\midrule
MapTRv2~\cite{liao2025maptrv2} & 20.0 & --  & 14.4 & 11.6 & 16.9 & 42.9 & 321.6 & 363.2 & 126.5 & 811.3 \\
Ours (cross-attention)   & 1.2  & 4   & 14.4 & 1.9  & 8.9  & 25.1 & 321.6 & 22.8  & 28.7  & 373.1  \\
Ours (self-attention)  & 1.2 & 4 & 14.4 & 1.7 & 8.9 & 24.9 & 321.6 & 33.1 & 28.7 & 383.4 \\
\midrule
\multicolumn{11}{c}{\textit{ResNet-18 backbone}} \\
\midrule
MapTRv2~\cite{liao2025maptrv2} & 20.0 & --  & 12.8 & 11.6 & 16.9 & 41.3 & 354.7 & 363.2 & 126.5 & 844.4 \\
Ours (cross-attention)   & 1.2  & 4   & 12.8 & 1.9  & 8.9  & 23.5 & 354.7 & 22.8  & 28.7  & 406.2  \\
Ours (self-attention)  & 1.2 & 4 & 12.8 & 1.7 & 8.9 & 23.3 & 354.7 & 33.1 & 28.7 & 416.5 \\
\bottomrule
\end{tabular}
}
\end{table}

\clearpage

\section{Ablation Studies}
\label{app:ablation}

\noindent \textbf{Ablation on Token Count and Mapper Depth.} Table~\ref{tab:ablation} examines the effect of token count and mapper depth on accuracy. Increasing the number of tokens from 0.9k to 1.2k improves accuracy, but further increasing to 1.5k degrades it, possibly due to increasing redundancy among 1D map tokens. Varying the mapper depth shows a similar pattern: a depth of 3 is insufficient, while a depth of 5 slightly degrades accuracy, consistent with prior observations that transformers can exhibit diminishing returns as attention representations become increasingly redundant~\cite{zhou2021deepvit}. Our final configuration of 1.2k tokens and a mapper depth of 4 therefore offers the best accuracy among the settings tested, with FPS remaining nearly constant across all configurations.

\noindent \textbf{Ablation on Timesteps.} Table~\ref{tab:ablation} also shows that accuracy improves consistently up to 4 timesteps and slightly degrades at 6, making 4 timesteps the best-performing setting. This observation is consistent with MemFusionMap~\cite{song2025memfusionmap}, which conjectures that excessive temporal fusion dilutes the current-frame feature, and MambaMap~\cite{yang2025mambamap}, which conjectures that an excessively long memory window introduces redundancy and noise.

\begin{table}[ht]
\centering
\caption{\textbf{Ablations over token count, mapper depth, and timesteps}. Our final configuration of 1.2k tokens, 4 mapper layers, and 4 timesteps (current frame + 3 past frames) achieves the best accuracy across configurations. FPS varies only from 18.8 to 19.2 across all configurations and is thus omitted.}
\vspace{1mm}
\label{tab:ablation}
\adjustbox{max width=\linewidth}{
\begin{tabular}{lcccc}
\toprule
Config & AP$_\text{div}$ & AP$_\text{ped}$ & AP$_\text{bound}$ & mAP$\uparrow$ \\
\midrule
\multicolumn{5}{c}{\textit{Ablation on token count (4 layers, 4 timesteps)}} \\
\midrule
0.9k tokens & 26.5 & 39.2 & 34.1 & 33.3 \\
\textbf{1.2k tokens} & \textbf{28.3} & \textbf{41.0} & \textbf{36.7} & \textbf{35.3} \\
1.5k tokens & 27.8 & 40.7 & 36.1 & 34.9 \\
\midrule
\multicolumn{5}{c}{\textit{Ablation on mapper depth (1.2k tokens, 4 timesteps)}} \\
\midrule
3 layers & 26.8 & 39.8 & 34.8 & 33.8 \\
\textbf{4 layers} & \textbf{28.3} & \textbf{41.0} & \textbf{36.7} & \textbf{35.3} \\
5 layers & 27.5 & 40.9 & 35.8 & 34.7 \\
\midrule
\multicolumn{5}{c}{\textit{Ablation on timesteps (4 layers, 0.3k tokens/step)}} \\
\midrule
1 timestep  & 23.1 & 35.1 & 32.2 & 30.1 \\
2 timesteps & 26.9 & 38.3 & 33.5 & 32.9 \\
\textbf{4 timesteps} & \textbf{28.3} & \textbf{41.0} & \textbf{36.7} & \textbf{35.3} \\
6 timesteps & 27.1 & 39.4 & 34.6 & 33.7 \\
\bottomrule
\end{tabular}
}
\end{table}

\clearpage

\section{Justification for the Near-Extrapolation Split}
\label{app:map_split}

\begin{figure}[ht]
    \centering
    \includegraphics[width=\linewidth]{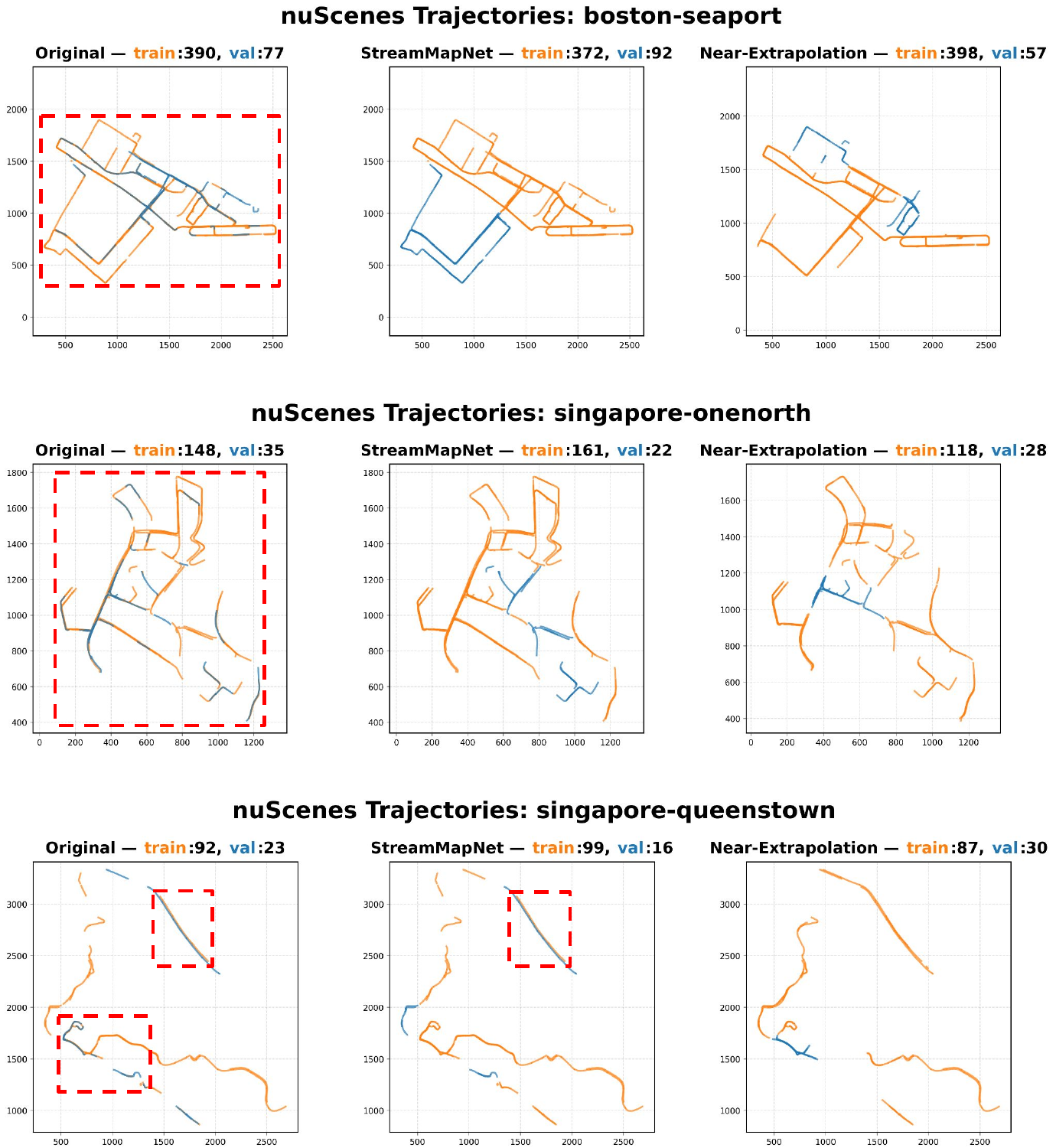}
    \caption{\textbf{Visualization of train--val geographic overlap across nuScenes regions.} The Near-Extrapolation split~\cite{lilja2024localization}, used in our online vectorized HD map construction experiments as the default setting, shows almost no train--val overlap. In contrast, the original split exhibits substantial train–val geographic overlap (79.4\%), and the StreamMapNet split retains residual overlap (2.1\%), indicating data leakage that undermines their reliability for evaluating online vectorized HD map construction. \textbf{\txtred{Red}} dashed boxes highlight regions of train--val geographic overlap within 5~m. Best viewed zoomed in.}
    \label{fig:map_split}
\end{figure}
\clearpage

\section{Robustness to Camera Extrinsic Perturbations} 
\label{app:robustness}

\begin{figure}[ht]
    \centering
    \includegraphics[width=0.48\linewidth]{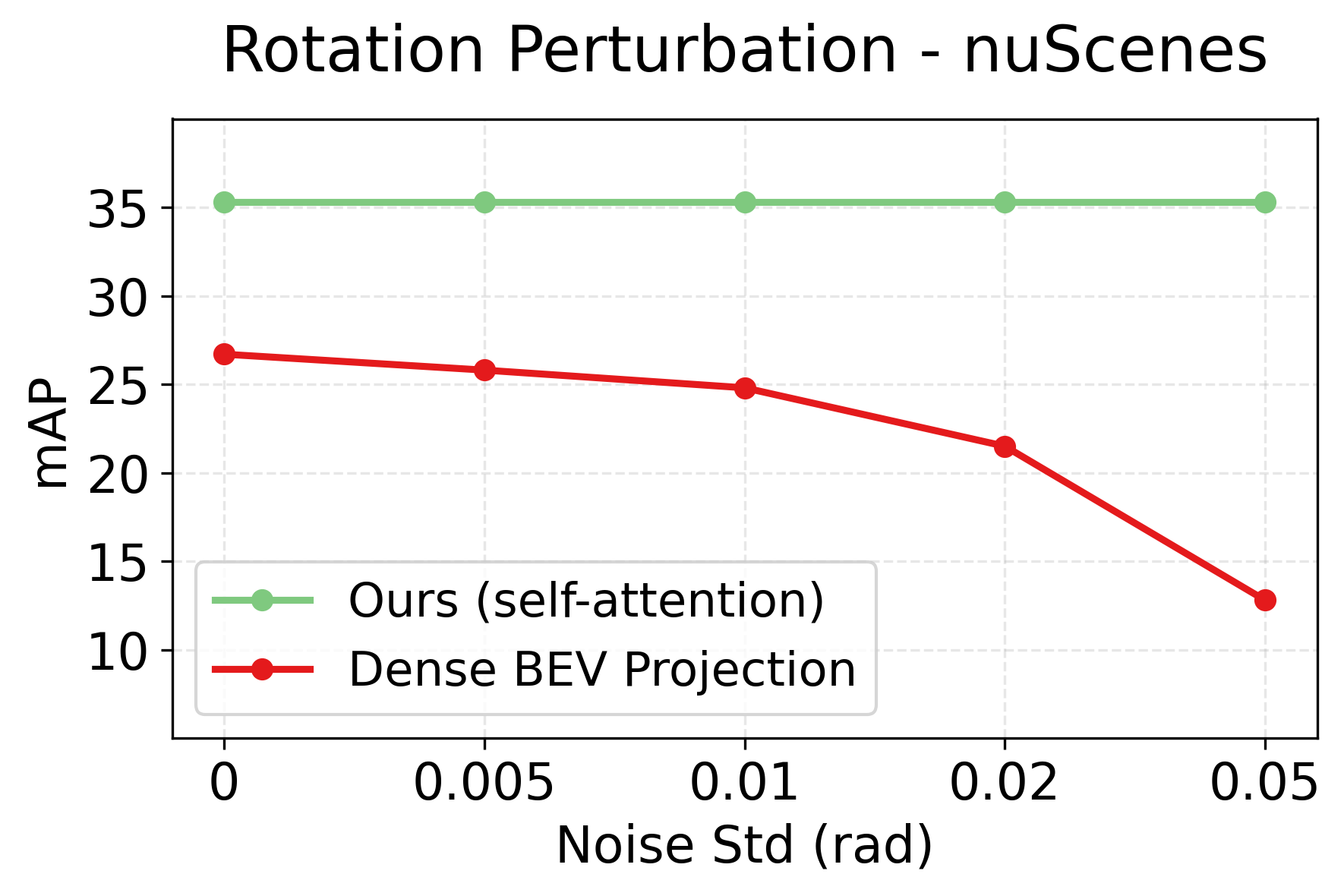}
    \includegraphics[width=0.48\linewidth]{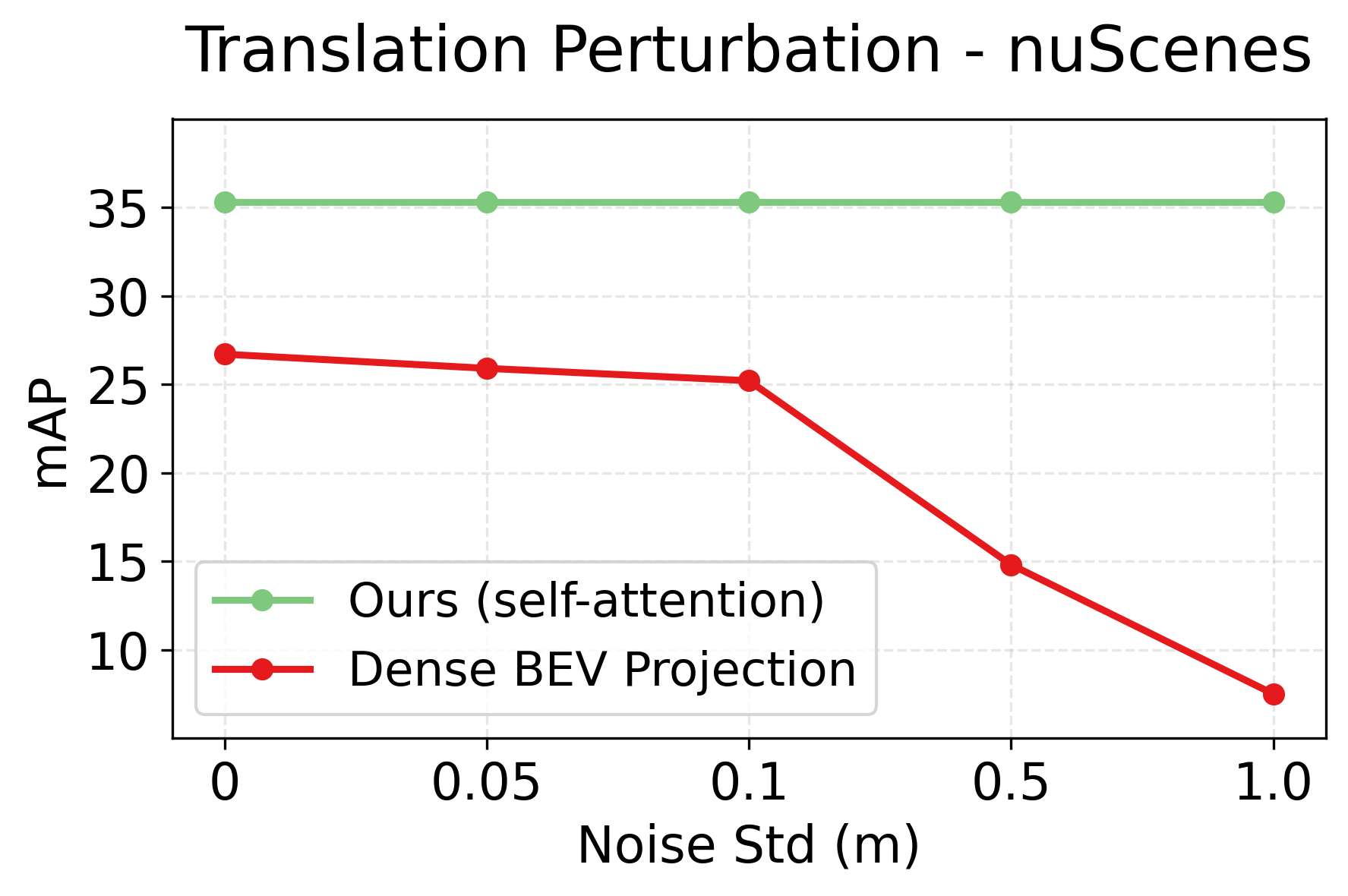}
     \includegraphics[width=0.48\linewidth]{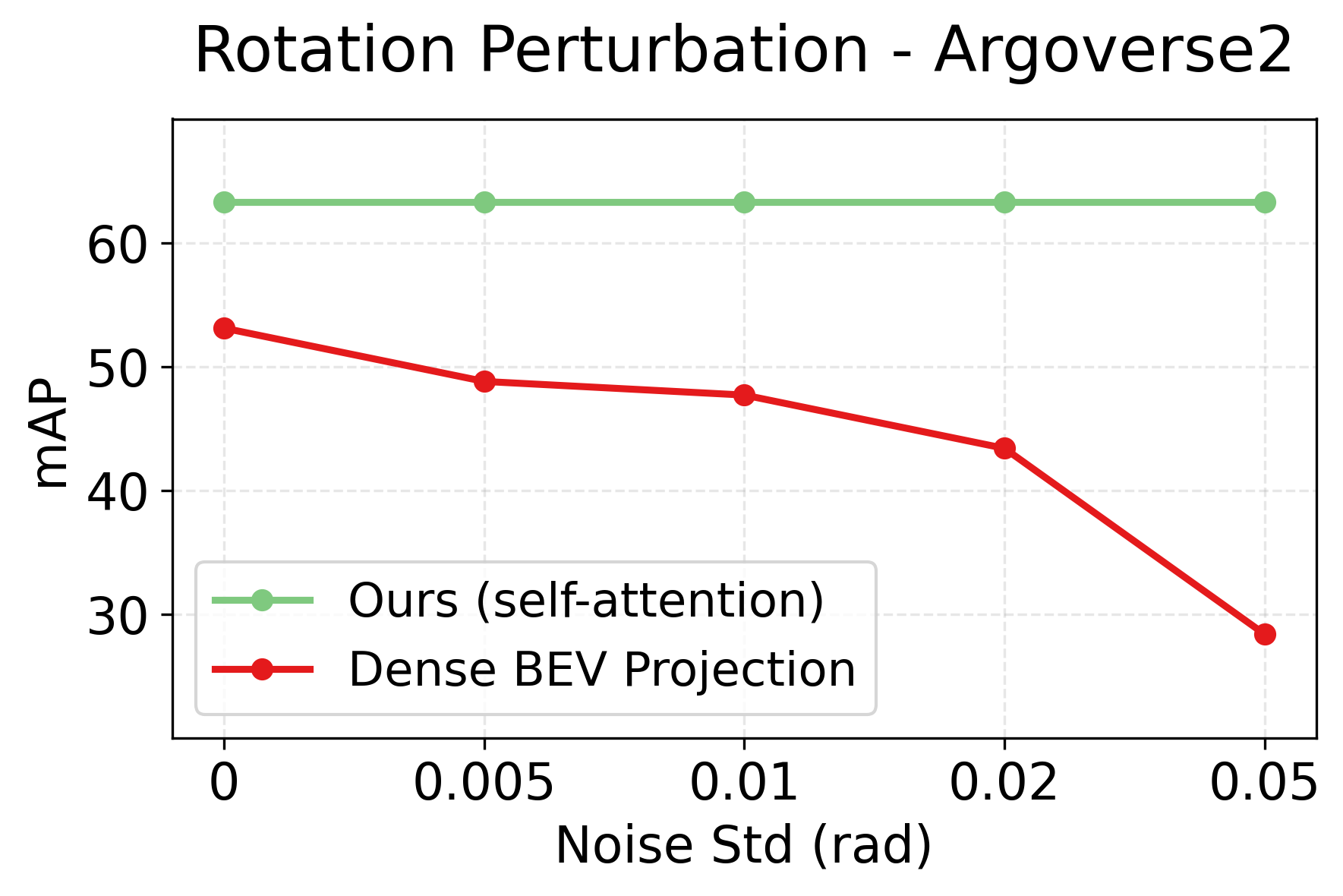}
      \includegraphics[width=0.48\linewidth]{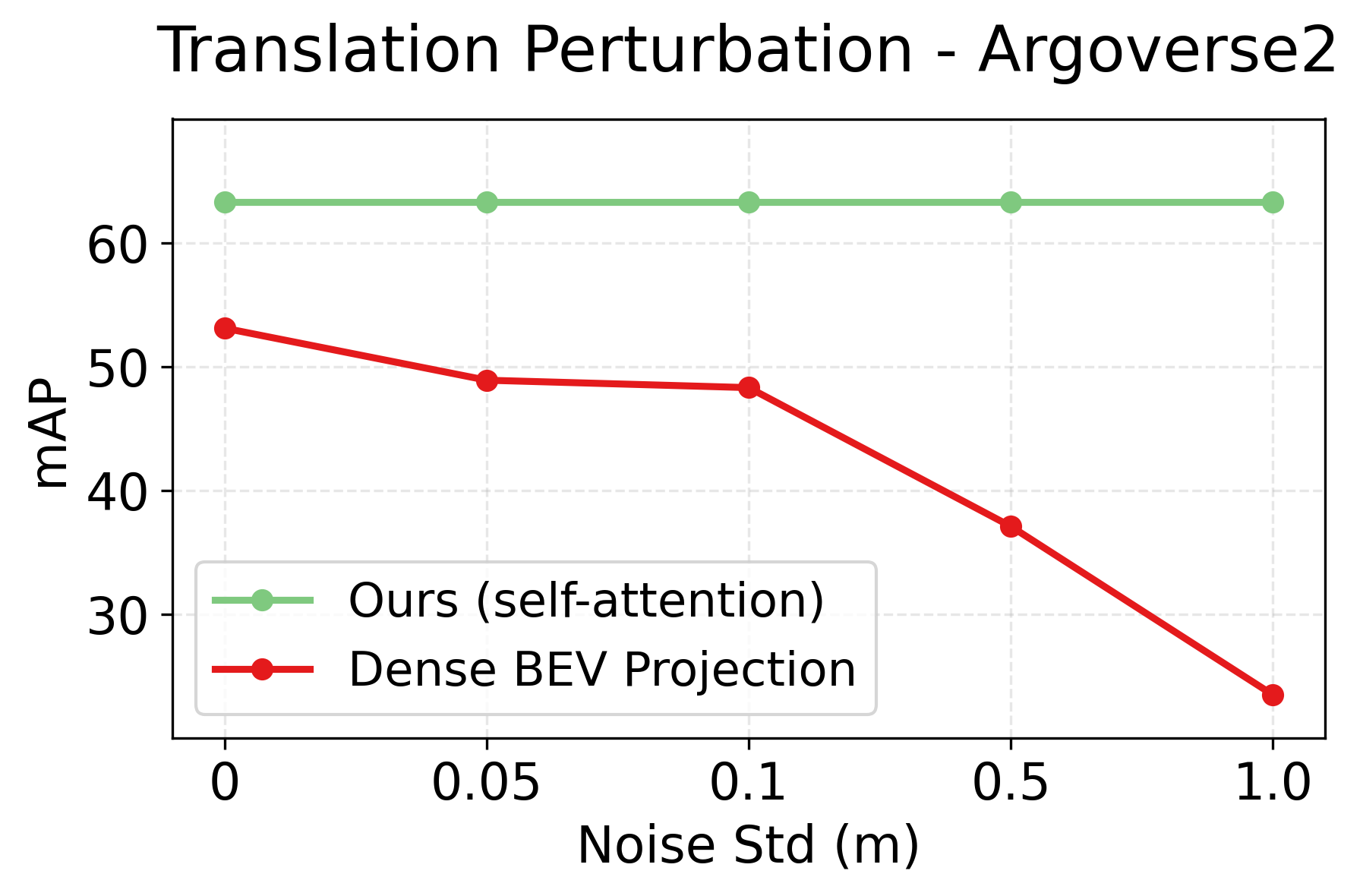}
    \caption{\textbf{Robustness to camera extrinsic perturbations.} We independently add zero-mean Gaussian noise to the rotation $(\theta_x,\theta_y,\theta_z)$ and translation $(\Delta x,\Delta y,\Delta z)$ components of all camera extrinsics. Our model does not use camera extrinsics and therefore shows no degradation, whereas Dense BEV Projection~\cite{liao2025maptrv2} explicitly relies on them and suffers substantial mAP drops under increasing perturbations.}
    \label{fig:camera_deviation}
\end{figure}

In real-world deployment, camera intrinsics are generally stable, while camera extrinsic parameters may be affected by calibration errors~\cite{liao2025maptrv2, wang2025efficient}. Following MapTRv2~\cite{liao2025maptrv2}, we evaluate robustness to camera extrinsic perturbations by independently adding zero-mean Gaussian noise with varying standard deviations to the rotation and translation components of the camera extrinsics. Results are reported in Figure~\ref{fig:camera_deviation}, where our method shows no degradation, maintaining 35.3 mAP on nuScenes and 63.3 mAP on Argoverse 2 across all perturbation levels, as it does not explicitly rely on camera extrinsics. In contrast, under the largest perturbations, Dense BEV Projection drops from 26.7 mAP to 12.8 under rotation and 7.5 under translation on nuScenes, and from 53.1 mAP to 28.4 under rotation and 23.5 under translation on Argoverse 2.

\clearpage

\section{Implementation Details}
\label{app:implement}

For both online vectorized HD map construction and 3D object detection, we replace the original BEV projection modules with our self-attention mapper and use full instead of deformable cross-attention in the decoder. Unlike the single-frame MapTRv2~\cite{liao2025maptrv2} setting, we use $T=4$ timesteps and 50 map tokens per camera per timestep, yielding 1.2k map tokens for nuScenes (6 cameras) and 1.4k for Argoverse~2 (7 cameras).

\subsection{Online Vectorized HD Map Construction}

We mainly follow the MapTRv2~\cite{liao2025maptrv2} framework and training settings, with ImageNet-pretrained~\cite{deng2009imagenet} image backbones and color-jitter augmentation. Input images are resized by a factor of 0.5 on nuScenes (1600$\times$900 to 800$\times$450) and, after padding to 2048$\times$2048, by a factor of 0.3 on Argoverse~2. The decoder uses 50 instance queries, 20 point queries per element, and 6 layers, with classification, point, and direction loss weights of 2, 5, and 0.005. We remove MapTRv2's depth, BEV segmentation, and PV segmentation auxiliary losses, since our mapper involves no depth estimation or BEV grid, and retain its one-to-many matching branch (300 extra instance queries, ground truth repeated 6 times). We train with AdamW~\cite{loshchilov2017decoupled} with a weight decay of 0.01 and a batch size of 4 per GPU on 8 NVIDIA RTX 3090 GPUs for 24 epochs on nuScenes and 6 epochs on Argoverse~2, with an initial learning rate of $6 \times 10^{-4}$ and cosine decay. Adjacent input frames are 0.5~s apart on nuScenes and 0.4~s apart on Argoverse~2.

\subsection{3D Object Detection}


We mainly follow the BEVFormer v2~\cite{yang2023bevformer} framework and training settings on nuScenes, including COCO-pretrained~\cite{lin2014microsoft} backbone initialization and image-level data augmentation, with a ResNet-50 backbone and a 6-layer detection decoder. Unlike BEVFormer v2, which also fuses future frames and uses first-stage perspective proposals as hybrid object queries, we use only past frames (4-frame input, adjacent frames 0.5~s apart) and learned object queries. For easier comparison with prior work~\cite{li2023bevdepth, lin2023sparse4d, wang2023exploring, liu2023sparsebev,han2024exploring,li2024bevnext,su2025freqpde,fan2026stur3d}, we use an input resolution of 256$\times$704. We train for 24 epochs on 8 NVIDIA RTX 3090 GPUs with a batch size of 2 per GPU using AdamW~\cite{loshchilov2017decoupled} with a weight decay of 0.01, an initial learning rate of $4 \times 10^{-4}$, and an image-backbone learning rate of $2 \times 10^{-4}$.

\end{document}